\pdfoutput=1

\documentclass[11pt]{article}

\usepackage[]{ACL2023}

\usepackage{times}
\usepackage{latexsym}

\usepackage[T1]{fontenc}

\usepackage[utf8]{inputenc}

\usepackage{microtype}

\usepackage{inconsolata}

\usepackage{amsmath}
\usepackage{graphicx}
\usepackage{amssymb}
\usepackage{longtable}
\usepackage{booktabs}
\usepackage{multirow}
\usepackage{makecell}
\usepackage{CJKutf8}
\usepackage{varwidth}
\usepackage{tikz}
\usepackage{mathrsfs}
\usepackage{ragged2e}
\usepackage{setspace}
\usepackage{caption}
\usepackage{overpic}
\usepackage{textcomp}
\usepackage{wasysym}
\usepackage{adjustbox}
\usepackage{kotex}
\usepackage[russian,english]{babel}
\usepackage[ruled,vlined]{algorithm2e}

\CJKfamily{bsmi}

\usepackage{xcolor}

\title{Finding and Editing Multi-Modal Neurons in Pre-Trained Transformers}

\newcommand*{\affaddr}[1]{#1}
\newcommand*{\affmark}[1][*]{\textsuperscript{#1}}
\newcommand*{\email}[1]{\texttt{#1}}

\author{Haowen Pan\affmark[\textnormal{1}], Yixin Cao\affmark[\textnormal{2}], Xiaozhi Wang\affmark[\textnormal{3}], Xun Yang\affmark[\textnormal{1}]\thanks{*Corresponding author.}, Meng Wang\affmark[\textnormal{4}] \\
\affaddr{\affmark[1]University of Science and Technology of China} \\
\affaddr{\affmark[2]School of Computer Science, Fudan University} \\
\affaddr{\affmark[3]Tsinghua University} \\
\affaddr{\affmark[4]Hefei University of Technology} \\
\email{phw1129@mail.ustc.edu.cn, caoyixin2011@gmail.com} \\
\email{wangxz20@mails.tsinghua.edu.cn, xyang21@ustc.edu.cn, wangmeng@hfut.edu.cn} \\
}

\begin{document}
\maketitle
\begin{abstract}
Understanding the internal mechanisms by which multi-modal large language models (LLMs) interpret different modalities and integrate cross-modal representations is becoming increasingly critical for continuous improvements in both academia and industry. In this paper, we propose a novel method to identify key neurons for interpretability --- how multi-modal LLMs bridge visual and textual concepts for captioning. Our method improves conventional works upon efficiency and applied range by removing needs of costly gradient computation. Based on those identified neurons, we further design a multi-modal knowledge editing method, beneficial to mitigate sensitive words or hallucination. For rationale of our design, we provide theoretical assumption. For empirical evaluation, we have conducted extensive quantitative and qualitative experiments. The results not only validate the effectiveness of our methods, but also offer insightful findings that highlight three key properties of multi-modal neurons: sensitivity, specificity and causal-effect, to shed light for future research.\footnote{We release our code at \url{https://github.com/opanhw/MM_Neurons}.}
\end{abstract}

\section{Introduction}
\label{sec:intro}

Recently, large language models (LLMs) have received much attention and become foundation models in many natural language processing applications~\citep{touvron2023llama, alpaca, vicuna2023, koala_blogpost_2023}. Following the success, researchers in the area of computer vision have extended the input modality to both text and image, namely multi-modal LLMs, showing remarkable performance in various visual understanding tasks~\citep{liu2023visual, instructblip, ye2023mplug, ye2023mplugowl2}. However, the underlying mechanism of how multi-modal LLMs interpret different modalities of features beyond these tasks remains unclear. It hinders in-depth investigation and poses risks in model applications, such as producing misleading outputs without insight into decisions or propagating biases through automatic captions.


\begin{figure*}
    \centering
    \begin{varwidth}[t]{\textwidth}
    \vspace*{0pt}
    \includegraphics[width=0.77\linewidth]{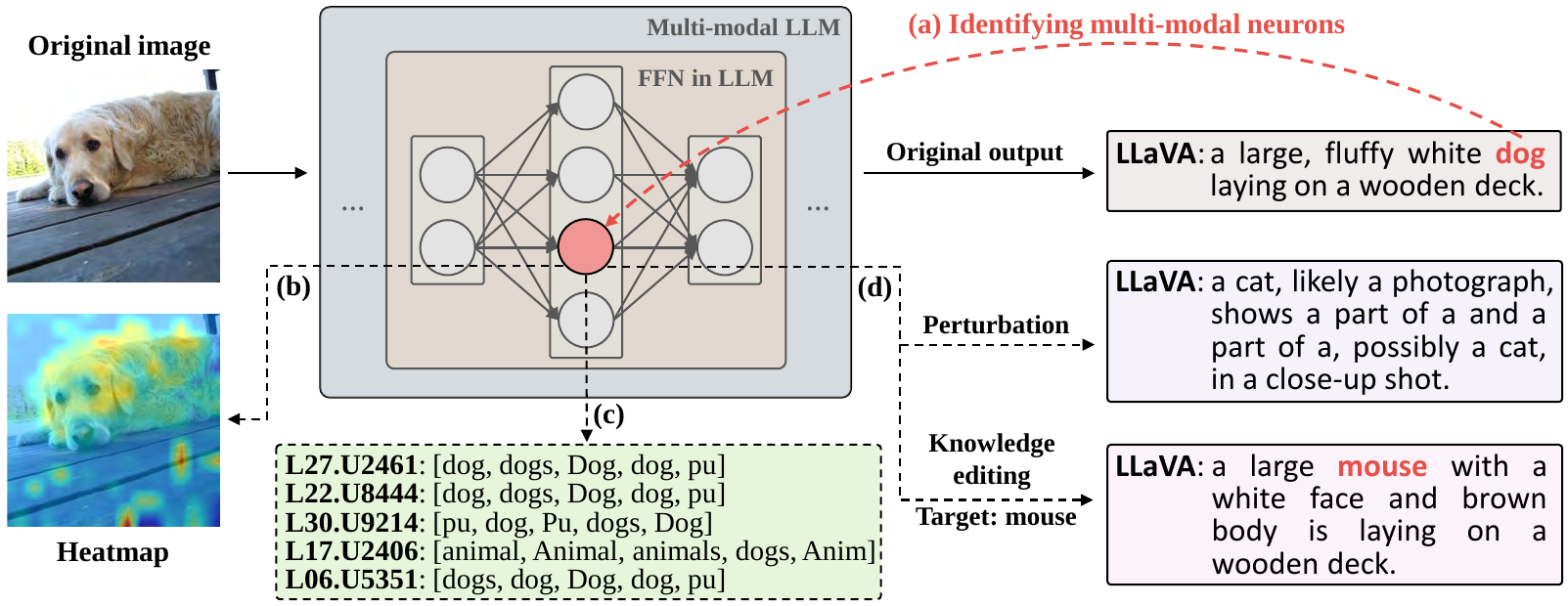} \\[-19.5pt] \begin{center}
        (\romannumeral1)
    \end{center}
    \end{varwidth}
    \hspace{5pt}
    \begin{varwidth}[t]{\textwidth}
    \centering
    \vspace*{9pt}
    \includegraphics[width=0.2\linewidth]{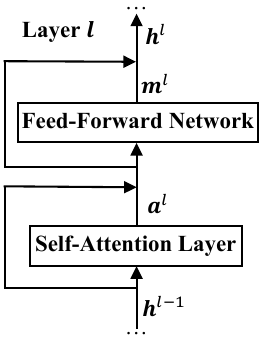} \\ \vspace{-1pt} \begin{center}
        (\romannumeral2)
    \end{center}
    \end{varwidth}
    \caption{(\romannumeral1) Multi-modal neurons in FFN within multi-modal LLM. We develop a method to \textbf{(a)} identify multi-modal neurons and confirm that they can encode specific concepts from \textbf{(b)} images to \textbf{(c)} texts and \textbf{(d)} causally affect model output. (\romannumeral2) Architecture of layer $l$ in Transformer-based LLM.}
    \label{fig:example_946}
\end{figure*}

There are two main types of methods on LLMs' interpretability. The first group targets probing various abilities through well-designed external tasks~\citep{olsson2022context, merullo2023language, huang2023look, duan2023shifting}. Another line of works, instead, attempt to reveal the internal states, by finding the processes of how LLMs understand and interpret textual inputs to form a response~\citep{meng2022locating, meng2022mass, dai2022kn, merullo2023language}. Among them, an interesting finding shows that LLMs' ability to understand textual information mainly comes from feed-forward networks (FFNs). Furthermore, \citet{schwettmann2023multimodal} identify key neurons from FFNs, namely multi-modal neurons. These neurons play an important role in understanding images and generating textual descriptions. However, the identification process is inefficient and limited in applied range, due to costly gradient computation. Besides, their theoretical rationale, empirical characteristics, and potential application remains under-exploration.




To address the issues, we propose a novel method for multi-modal neurons identification. We define a contribution score based on the activation output in FFNs, which is consistent with the probability distribution when predicting. As our method do not need access to the model gradients, we improve efficiency while ensuring effectiveness.

Based on the identified neurons, we further propose a multi-modal knowledge editing method as a potential application. We achieve the goal of editing a specific concept to another designative concept (e.g., in Figure \ref{fig:example_946}(\romannumeral1), `dog' is edited to `mouse'), by changing the probability distribution of outputs. Without additionally training the entire model or requiring access to model gradients, our proposed method facilitates a timely and resource-efficient editing of a small portion of the model parameters.

For empirical characteristics, we have designed metrics and conducted extensive experiments, which highlight three critical properties of multi-modal neurons: (1) \textbf{Sensitivity} (\S \ref{subsec:sensitivity}). Multi-modal neurons are sensitive to particular concepts. Once they are activated by some regions of the input image, they are responsible for generating related textual concepts. More importantly, these neurons are invariant in visual translation to different inputs. (2) \textbf{Specificity} (\S \ref{subsec:specificity}). Although different multi-modal neurons can be activated by the same concepts, they are selectively active for these concepts and hardly respond to others. (3) \textbf{Causal-Effect} (\S \ref{subsec:causal_effect}). Multi-modal neurons and the associated concepts have causal-effect and are significantly susceptible. We perturb and edit the identified multi-modal neurons, which leads to significant changes in outputs.

Our contributions can be summarized as follows:
\begin{itemize}
    \item We propose a new method for identifying multi-modal neurons in Transformer-based multi-modal LLMs.
    \item We propose a multi-modal knowledge editing method based on the multi-modal neurons.
    \item We highlight three critical properties of multi-modal neurons by designing four quantitative evaluation metrics and extensive experiments.
\end{itemize}

\section{Method}
\label{sec:method}
We first define neurons in the LLM (\S \ref{subsec:neurons_In_transformer}), and then define a contribution score for neurons identification (\S \ref{subsec:finding_vision_language_skill_neurons}). Furthermore, we propose a multi-modal knowledge editing method based on identified neurons (\S \ref{subsec:targeted_modification}) and introduce several evaluation metrics to evaluate multi-modal neurons (\S \ref{subsec:evaluation_metrics}).


\subsection{Neurons in Transformer-Based LLM}
\label{subsec:neurons_In_transformer}
A multi-modal LLM typically consists of an image encoder, a textual LLM, and an adaptor to align the above two modules. Following previous works~\citep{dai2022kn, wang-wen-etal2022skill, schwettmann2023multimodal}, we research neurons within FFNs in textual LLM, as they carry two-thirds of the parameters and are proven to play a critical role in understanding textual and visual features. Layers within a Transformer-based~\citep{vaswani2017attention} LLM can be illustrated as Figure \ref{fig:example_946}(\romannumeral2), where we denote the hidden states at layer $l$ as $\mathbf{h}^l$, the FFN output as $\mathbf{m}^l$ and the self-attention output as $\mathbf{a}^l$, respectively. And $\mathbf{m}^l$ can be calculated by:
\begin{align}
\label{eq:formula1}
\mathbf{m}^l = \mathbf{W}_{\text{out}}^{l}~\sigma\left(\mathbf{W}_{\text{in}}^{l} \left(\mathbf{a}^l + \mathbf{h}^{l- 1}\right)\right)~,
\end{align}
where $\mathbf{h}^{0}$ is the embedding vector of input, $\sigma$ is an activation function, $\mathbf{W}_{\text{in}}^{l}$ is the first linear layer and $\mathbf{W}_{\text{out}}^{l}$ is the second linear layer in FFN. And we omit the normalization in Eq.~\ref{eq:formula1} for the sake of brevity.

For simplicity, let $\mathbf{O}^l = \sigma\left(\mathbf{W}_{\text{in}}^{l} \left(\mathbf{a}^l + \mathbf{h}^{l-1}\right)\right)$, where the $i$-th element is the activation output of the $i$-th neuron. We denote each neuron in the LLM as (L$l$.U$i$) in subsequent experiments. For instance, (L20.U188) denotes the 188-th neuron at layer 20.

\subsection{Identifying Multi-Modal Neurons}
\label{subsec:finding_vision_language_skill_neurons}
We now propose a contribution score that indicates a neuron's contribution to a modal-independent concept. That is, if the score is high, the neuron should be activated with a high probability when taking in the visual concept and generating the textual concept.
We first formally define the computational method for it and then prove its validity.

Let $\mathcal{M}$ be the LLM, $\mathbf{x}$ be the sequence of input tokens and $\mathbf{y}$ be the output sequence. The function of LLM can be written as: $\mathbf{y} = \mathcal{M}(\mathbf{x})$.

We assume the model is about to output token $t\in \mathbf{y}$, whose probability is maximum among the vocabulary. Then we define the contribution score of the neuron $u_i$ at layer $l$ to the token $t$ as $s^l_{i, t}$:
\begin{align}
\label{eq:formula9}
s^l_{i, t} = \mathbf{Q}^l(i, t)~,
\end{align}
where $\mathbf{Q}^l = \mathbf{W}_u\mathbf{W}_{\text{out}}^{l}\circ\mathcal{T}\left(\mathbf{O}^l_{-1}\right) \in \mathbb{R}^{d_m\times v}$, $\mathbf{W}_u$ is the unembedding matrix to decode last hidden states, $\mathcal{T}\left(\cdot\right)$ is the transpose of the input matrix, $\mathbf{O}^l_{-1}$ is activation output at the last token, $d_m$ is intermediate size, $v$ is vocab size and $\circ$ is an element-wise product with broadcasting mechanism.

To validate rationality and effectiveness of Eq.~\ref{eq:formula9} and explain why we define $\mathbf{Q}^l$ in the manner described above, we try to disassemble and deduce the generation procedure of LLM. When a $L$ layer LLM is generating a new token $t\in \mathbf{y}$, the probability distribution of output can be denoted as follows:
\begin{align}
\label{eq:formula7}
t &= \operatorname{arg max}\left(\mathbf{W}_u\mathbf{h}^L_{-1}\right) \notag \\
&= \operatorname{arg max}\left(\mathbf{W}_u\left(\mathbf{a}^L_{-1}+\mathbf{m}^L_{-1}+\mathbf{h}^{L-1}_{-1}\right)\right) \notag \\
&= \operatorname{arg max}\left(\sum_{l=1}^{L}\left(\mathbf{W}_u\mathbf{m}^l_{-1} + \mathbf{W}_u\mathbf{a}^l_{-1}\right)\right. \notag \\
&\phantom{= \operatorname{arg max}\Bigg(} + \mathbf{W}_u\mathbf{h}^0_{-1}\Bigg) \notag \\
&= \operatorname{arg max}\left(\sum_{l=1}^{L}\left(\mathbf{W}_u\mathbf{W}_{\text{out}}^{l}\mathbf{O}^l_{-1} + \mathbf{W}_u\mathbf{a}^l_{-1}\right)\right. \notag \\
&\phantom{= \operatorname{arg max}\Bigg(} + \mathbf{W}_u\mathbf{h}^0_{-1}\Bigg)~,
\end{align}
where $\mathbf{W}_u$ is the unembedding matrix, $\mathbf{h}^L_{-1}$ is the output of the last token at the last layer $L$, and $\mathbf{O}^l_{-1} = \sigma\left(\mathbf{W}_{\text{in}}^{l} \left(\mathbf{a}^l_{-1} + \mathbf{h}^{l-1}_{-1}\right)\right) \in \mathbb{R}^{d_m}$ is activation function output at the last token at layer $l$.

In Eq.~\ref{eq:formula7}, $\mathbf{W}_u\mathbf{W}_{\text{out}}^{l}\mathbf{O}^l_{-1}$ represents FFN part and $\mathbf{W}_u\mathbf{a}^l_{-1}$ represents self-attention part. Following \S \ref{subsec:neurons_In_transformer}, we empirically focus on the FFN and omit the remaining parts. We regard $o_i^l$, the $i$-th element of $\mathbf{O}^l_{-1}$, as the activation of the $i$-th neuron at the last token at layer $l$, and $\mathbf{W}_u\mathbf{W}_{\text{out}}^{l}$ as a new unembedding matrix at each layer. The function of $\mathbf{W}_u\mathbf{W}_{\text{out}}^{l}$ is to project the activation of the neurons onto a distribution of the token vocabulary. The distributions at each layer then are summed up to obtain a final distribution, containing contributions of all neurons within the model.

To further evaluate the individual contribution of each neuron, we disassemble the matrix multiplication of $\mathbf{W}_u\mathbf{W}_{\text{out}}^{l}$ and $\mathbf{O}^l_{-1}$ in Eq.~\ref{eq:formula7} as follows:
\begin{align}
\label{eq:formula8}
\textstyle{\mathbf{W}_u\mathbf{W}_{\text{out}}^{l}\mathbf{O}^l_{-1} = \sum \mathcal{T}\left(\mathbf{W}_u\mathbf{W}_{\text{out}}^{l}\circ\mathcal{T}\left(\mathbf{O}^l_{-1}\right)\right),}
\end{align}
where $\sum\left(\cdot\right)$ represents summing rows of the input.

Now we can see $\mathbf{Q}^l$ in Eq.~\ref{eq:formula8}, which is consistent with the probability distribution when predicting. We regard $\mathbf{Q}^l(i, j)$ as a contribution score that the $i$-th neuron at layer $l$ contributes to the $j$-th token. We provide a more detailed explanation in Appendix~\ref{sec:supplementary_explanation}.

Based on Eq. \ref{eq:formula9}, we compute the score of each neuron for every \textbf{noun} token in the model output. Then we rank all scores of neurons across all layers within the model by the descending order and regard the top neurons as multi-modal neurons.
Implementation details can be found in Appendix \ref{subsec:identifying}.

\begin{algorithm}[t]
    \small
    \SetAlgoLined
    \KwData{Source token $t_0$, target token $t_1$, neurons set $\mathcal{S}$, model $\mathcal{M}$, unembedding matrix $\mathbf{W}_u$, penalty weight $\beta$, learning rate $\alpha$, epochs $\epsilon$}
    \KwResult{Edited model $\tilde{\mathcal{M}}$}
    \nl \For{$s_j \in \mathcal{S}$}{
    \nl   $l, i \gets \text{location of~} s_j$\;
    \nl   $o^l_i \gets \text{activation function output of~} s_j$\;
    \nl   $\mathbf{w} \gets i\text{-th row of~} \mathbf{W}_{\text{out}}^{l}$\;
    \nl   $\mathbf{v}_0 \gets t_0\text{-th column of~} \mathbf{W}_u$\;
    \nl   $\mathbf{v}_1 \gets t_1\text{-th column of~} \mathbf{W}_u$\;
    \nl   initialize $\Delta\mathbf{w}$\;
    \nl   $\mathbf{w}' \gets \mathbf{w} + \Delta\mathbf{w}$\;
    \nl   $\text{loss~} \gets o^l_i(\mathbf{w}'\mathbf{v}_0 - \mathbf{w}'\mathbf{v}_1) + \beta \cdot ||\Delta\mathbf{w}||_2$\;
    \nl   $\Delta\mathbf{w}^* \gets \text{gradient descent}(\Delta\mathbf{w}, \text{loss}, \alpha, \epsilon)$\;
    \nl   $\tilde{\mathbf{W}}_{\text{out}}^{l} \gets \text{add~}\Delta\mathbf{w}^*\text{~to the~} i\text{-th row of~} \mathbf{W}_{\text{out}}^{l}$\;
    \nl   $\tilde{\mathcal{M}} \gets \text{replace~} \mathbf{W}_{\text{out}}^{l}\text{~with~}\tilde{\mathbf{W}}_{\text{out}}^{l}\text{~in~}{\mathcal{M}}$\;
    }
    \nl \Return $\tilde{\mathcal{M}}$;
\caption{Knowledge Editing}
\label{algo:targeted_modification}
\end{algorithm}

\subsection{Multi-Modal Knowledge Editing}
\label{subsec:targeted_modification}
Following previous works~\citep{mitchell2022fast, meng2022locating, meng2022mass} on unimodal knowledge editing, we aim at controlling the textual output. In specific, our goal is to replace a source token with a target token in the output without changing the remaining content.
We propose an algorithm (see Algorithm~\ref{algo:targeted_modification}) to intervene some parameters based on the identified multi-modal neurons.

We denote top multi-modal neurons of source token $t_0$ as $\mathcal{S}$. For each multi-modal neuron $s_j \in \mathcal{S}$, we first get its location $(l, i)$, which means the $i$-th neuron at layer $l$, and then we record its activation function output $o_i^l$. Let $\mathbf{w}$ be the $i$-th row of $\mathbf{W}_{\text{out}}^{l}$, $\mathbf{v}_0$ be the $t_0$-th column of $\mathbf{W}_u$, $\mathbf{v}_1$ be the $t_1$-th column of $\mathbf{W}_u$ and $\mathbf{w}'$ be the edited $\mathbf{w}$, respectively.

Our goal is to prompt the probability of generating token $t_1$ higher than token $t_0$, which is equivalent to make $o^l_i\mathbf{w}'\mathbf{v}_1$ larger than $o^l_i\mathbf{w}'\mathbf{v}_0$, so we define a loss function as below:
\begin{align}
\text{loss} = o^l_i(\mathbf{w}'\mathbf{v}_0 - \mathbf{w}'\mathbf{v}_1) + \beta \cdot ||\Delta\mathbf{w}||_2~,
\end{align}
where $\beta$ is penalty weight and $||\Delta\mathbf{w}||_2$ is a $L_2$-norm constraint as a penalty to avoid the editing is too drastic and affects generating other tokens.

By applying Gradient Descent~\citep{robbins1951stochastic}, we acquire an optimal $\Delta\mathbf{w}^*$. We then add $\Delta\mathbf{w}^*$ to the $i$-th row of $\mathbf{W}_{\text{out}}^{l}$ and replace the original $\mathbf{W}_{\text{out}}^{l}$ with the new $\mathbf{W}_{\text{out}}^{l}$ in model $\mathcal{M}$.

Note that our algorithm is independent from the model, and the solution procedure does not need to additionally train or infer the entire model. Accordingly, this allows for an efficient, timely and resource-efficient editing of the model parameters.

\subsection{Evaluation Metrics}
\label{subsec:evaluation_metrics}
After identifying multi-modal neurons, in order to comprehensively evaluate the effectiveness of them with quantitative indicators, we measure several evaluation metrics from multiple perspectives.

\noindent\textbf{Semantic Sensitivity:}
\quad To verify if neurons are sensitive to textual concepts, we align neurons with natural language. The more similar the top tokens are to the textual concept, the more sensitive the neurons are. Therefore, we measure BERTScore~\citep{zhang2019bertscore}, MoverScore~\citep{zhao2019moverscore} and BLEURT~\citep{sellam2020bleurt} between each textual concept and top-10 tokens that corresponding neurons represent.

\noindent\textbf{Region Invariance:}\quad To verify if neurons are sensitive to visual concepts, we measure the proportion of invariant neurons when shuffling the image patches.
Specifically, for each textual concept in each image, we denote the original top-$k$ multi-modal neurons as $\mathcal{S}_k$. We randomly shuffle the input sequence of image patches of LLM, and equally identify top-$k$ multi-modal neurons, denoted as $\mathcal{S}'_k$. A higher degree of similarity between $\mathcal{S}_k$ and $\mathcal{S}'_k$ indicates stronger region invariance. We calculate the ratio of invariant neurons as below:
\begin{align}
\label{eq:formula10}
r_k = \frac{|\mathcal{S}_k\cap \mathcal{S}'_k|}{|\mathcal{S}_k|}~,
\end{align}
and record a mean score across all images.

\noindent\textbf{Cross-Images Invariance:}\quad We aim at figuring out whether the same neurons would be identified in different images, which is called cross-images invariance. We randomly select $N$ different images from the dataset that all contain a given concept $c$. Then, we separately identify the top-$k$ neurons of these images and pick out neurons in common. We calculate the ratio of common neurons by:
\begin{align}
\label{eq:formula11}
    s_{\text{CII}} = \frac{|\mathcal{S}_k^1\cap\mathcal{S}_k^2\cap \cdots \cap \mathcal{S}_k^N|}{k}~,
\end{align}
where $\mathcal{S}_k^j$ is top-$k$ multi-modal neurons of image $j$.

\noindent\textbf{Specificity:}\quad 
We then verify if neurons are specific to textual concepts --- only activated for some related tokens, but inactivated for other tokens.
Formally, we pick out $n$ images, and separately identify their top-1 multi-modal neuron, denoted as $\mathcal{S}$. For each neuron $(l, i)$ in $\mathcal{S}$, we provide a set of concepts $T$, where $|T|=m$, and calculate scores to each of them. Then we record a mean score across neurons in $\mathcal{S}$ and concepts in $T$, denoted as S@$m$:
\begin{align}
\label{eq:formula12}
    \text{S@}m = \frac{1}{n\cdot m}\sum_{(l, i) \in \mathcal{S}}\sum_{t\in T} s^l_{i, t}~.
\end{align}

We choose two sets of concepts $T$: related concepts and random concepts. Related concepts are concepts with top probability to each neuron in $\mathcal{S}$, while random concepts are randomly selected from the vocabulary. If multi-modal neurons possess specificity, scores to related concepts will significantly outperform those to random concepts.

We measure semantic sensitivity in \S\ref{subsubsec:determining_textual_meanings}, region invariance in \S\ref{subsubsec:position_invariance}, cross-images invariance in \S\ref{subsubsec:cross_images_invariance} and specificity in \S\ref{subsec:specificity}, respectively.

\section{Experiments}
\label{sec:exp}
\subsection{Investigation Setup}
\label{subsec:investigation_setup}
We use LLaVA~\citep{liu2023visual}, InstructBLIP~\citep{instructblip} and mPLUG-Owl2~\citep{ye2023mplugowl2} as our research models, which are three widely-use models for visual semantic understanding task. And we conduct all experiments on 1000 images that are randomly sampled from SBU Captions Dataset~\citep{Ordonez:2011:im2text}, a dataset consists of more than 1 million images from Flickr. We compare our method with Multimodal Neurons (abbreviated as Mmns)~\citep{schwettmann2023multimodal}, a technique for detecting \textit{multimodal neurons} that map visual features to corresponding text. Furthermore, we establish a baseline (abbreviated as Base) that simply selects neurons with higher activations at the last token for basic comparison. Details about the implementations can be found in appendix \ref{subsec:identifying}.

\begin{figure}[t]
    \centering
    \includegraphics[width=\linewidth]{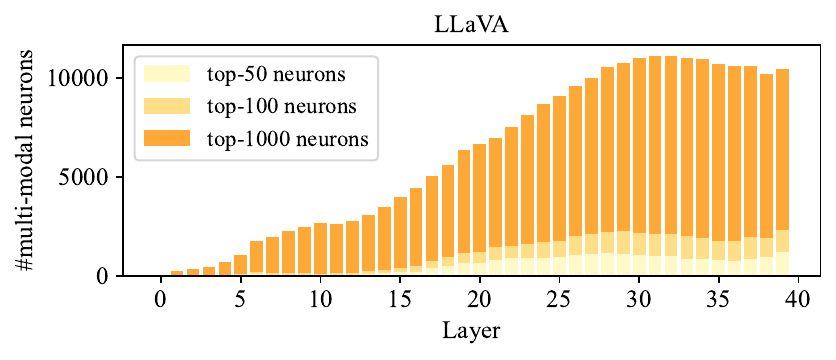}\\
    \includegraphics[width=\linewidth]{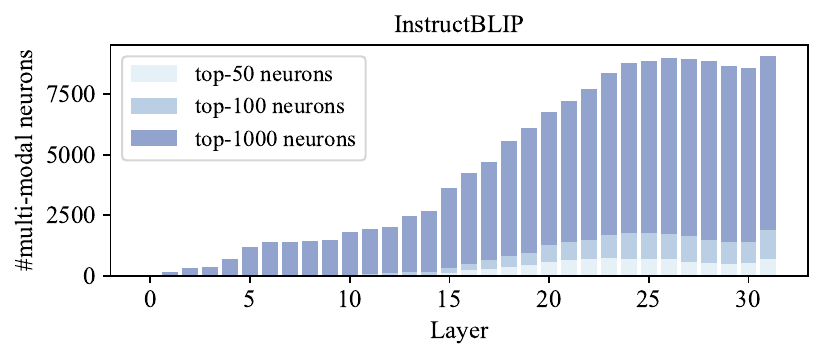}\\
    \includegraphics[width=\linewidth]{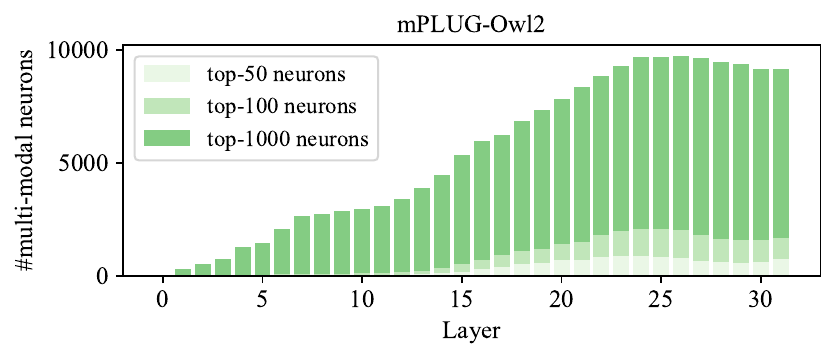}
    \caption{Distribution of unique multi-modal neurons per layer, chosen by different number of neurons with top contribution scores for each image.}
    \label{fig:distribution_k_50_100_1000}
\end{figure}

\begin{table}[t]
    \centering
    \renewcommand\arraystretch{0.7}
    \begin{tabular}{@{\hspace{4pt}}c@{\hspace{6pt}}c@{\hspace{4pt}}c@{\hspace{4pt}}c@{\hspace{4pt}}c@{\hspace{4pt}}}
        \toprule
        \multicolumn{5}{c}{\textbf{\small Image \& Original output}} \\
    \midrule
    \multicolumn{5}{c}{\includegraphics[width=0.22\linewidth]{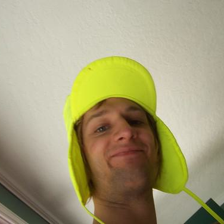} \hspace{5pt}\begin{varwidth}[f]{\textwidth}
        \vspace*{-40pt}\makecell[l]{\small \parbox{4cm}{LLaVA: a \textbf{man} wearing a yellow \phantom{LLaVA: }\textbf{hat} and smiling.}}\end{varwidth}} \\
        \midrule
        \multirow{2}{*}{\small \bf Concept} & \multicolumn{4}{c}{\small \bf Heatmap \& Binary mask} \\
        & \scriptsize \bf Top-1 & \scriptsize \bf Top-10 & \scriptsize \bf Top-100 & \scriptsize \bf Top-1000 \\
        \midrule
       \multirow{4.5}{*}{\raisebox{-0.0\height}{\small man}} &{\raisebox{-0\height}{\makecell{\includegraphics[width=0.08\textwidth]{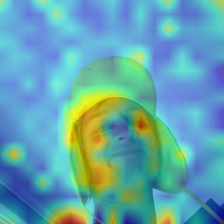}}}} & {\raisebox{-0\height}{\makecell{\includegraphics[width=0.08\textwidth]{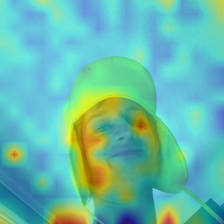}}}} & {\raisebox{-0\height}{\makecell{\includegraphics[width=0.08\textwidth]{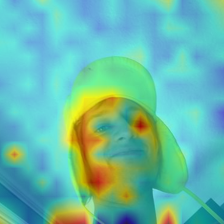}}}} & {\raisebox{-0\height}{\makecell{\includegraphics[width=0.08\textwidth]{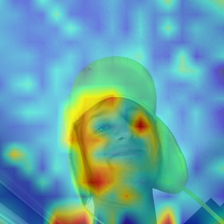}}}} \\
        & {\raisebox{-0\height}{\makecell{\includegraphics[width=0.08\textwidth]{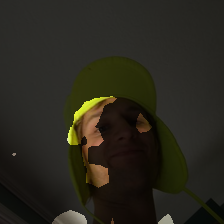}}}} & {\raisebox{-0\height}{\makecell{\includegraphics[width=0.08\textwidth]{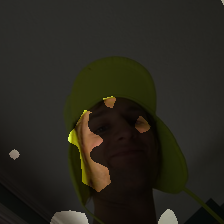}}}} & {\raisebox{-0\height}{\makecell{\includegraphics[width=0.08\textwidth]{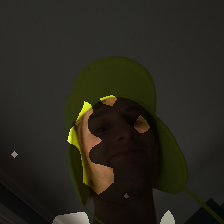}}}} & {\raisebox{-0\height}{\makecell{\includegraphics[width=0.08\textwidth]{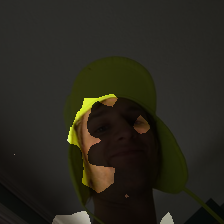}}}} \\[-2.5pt]
        
        \cmidrule(lr){1-5}

        \multirow{4.5}{*}{\raisebox{-0.0\height}{\small hat}} & {\raisebox{-0\height}{\makecell{\includegraphics[width=0.08\textwidth]{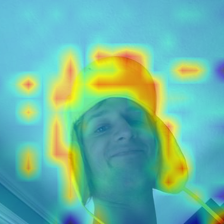}}}} & {\raisebox{-0\height}{\makecell{\includegraphics[width=0.08\textwidth]{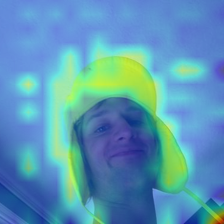}}}} & {\raisebox{-0\height}{\makecell{\includegraphics[width=0.08\textwidth]{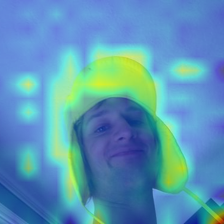}}}} & {\raisebox{-0\height}{\makecell{\includegraphics[width=0.08\textwidth]{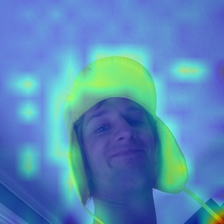}}}} \\
        & {\raisebox{-0\height}{\makecell{\includegraphics[width=0.08\textwidth]{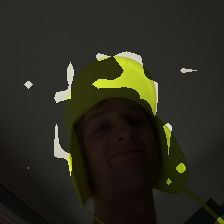}}}} & {\raisebox{-0\height}{\makecell{\includegraphics[width=0.08\textwidth]{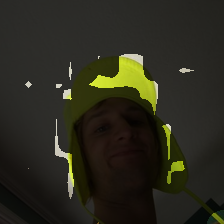}}}} & {\raisebox{-0\height}{\makecell{\includegraphics[width=0.08\textwidth]{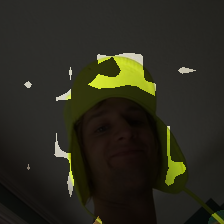}}}} & {\raisebox{-0\height}{\makecell{\includegraphics[width=0.08\textwidth]{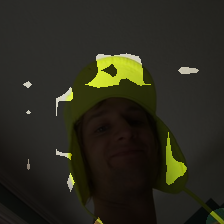}}}} \\[-2.5pt]
        \bottomrule
    \end{tabular}
    \caption{Heatmap and binary mask results of an example image. We plot each heatmap by using scaled mean activations across top-$k$ neurons, where $k=1, 10, 100, 1000$, and plot binary mask by thresholding mean activations above the 95\% percentile, respectively.}
    \label{tab:visual_example}
\end{table}

\begin{table*}[t]
    \small
    \centering
    \renewcommand\arraystretch{0.6}
    \begin{tabular}{cccll}
        \toprule
        \bf Image & \bf Model & \bf Method & \multicolumn{1}{c}{\bf Top neurons} & \multicolumn{1}{c}{\bf Top tokens} \\
        \midrule
        \multirow{1}{*}{\raisebox{-1.35\height}{\makecell{\includegraphics[width=0.15\textwidth]{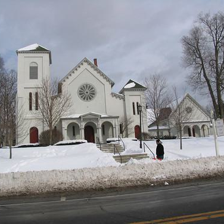} \\[4pt] \hspace{-2pt} \makecell[c]{\scriptsize \parbox{0.8cm}{\vspace*{-17.3pt} LLaVA: } \parbox{2.4cm}{a \textbf{church} with a steeple, surrounded by snow, is captured in the photo.}} \\[16pt] \hspace{-2pt} \makecell[c]{\scriptsize \parbox{1.35cm}{\vspace*{-9.2pt} InstructBLIP: } \parbox{1.9cm}{a \textbf{church} with snow on the ground.}} \\[12pt] \hspace{-2pt} \makecell[c]{\scriptsize \parbox{1.54cm}{\vspace*{-16.0pt} mPLUG-Owl2: } \parbox{1.9cm}{a \textbf{church} with a person shoveling snow in front of it.}}}}} & \multirow{11}{*}{\raisebox{-0.4\height}{LLaVA}} & \multirow{3}{*}{Base} & \scriptsize ~~ \bf L39.U212 & \scriptsize [`', `1', `-', `\textbackslash n', `('] \\
        & & & \scriptsize ~~ \bf L24.U5916 & \scriptsize [`arin', `Kennedy', `dy', `dy', `PF'] \\
        & & & \scriptsize ~~ \bf L39.U5925 & \scriptsize [` ', `----', `-----', `\_\_\_\_\_\_\_\_', `\_\_\_\_'] \\
        \cmidrule(lr){3-5}
        & & \multirow{3}{*}{Mmns} & \scriptsize ~~ \bf L24.U10906 & \scriptsize [`dex', `igung', `nomin', `pill', `pill'] \\
        & & & \scriptsize ~~ \bf L9.U4426 & \scriptsize [`,', `.', `bird', `bird', `-'] \\
        & & & \scriptsize ~~ \bf L20.U3864 & \scriptsize [`oka', `backwards', `\foreignlanguage{russian}{рем}', `iono', `차'] \\
        \cmidrule(lr){3-5}
        & & \multirow{3}{*}{Ours} & \scriptsize ~~ \bf L31.U9192 & \scriptsize [`church', `Church', `churches', `Kirche', `Kirchen'] \\
        & & & \scriptsize ~~ \bf L34.U8761 & \scriptsize [`religious', `Relig', `relig', `religion', `Catholic'] \\
        & & & \scriptsize ~~ \bf L39.U9669 & \scriptsize [`Church', `Luther', `Bishop', `Orth', `church'] \\
        
        \cmidrule(lr){2-5}
        
        & \multirow{11}{*}{\raisebox{-0.4\height}{InstructBLIP}} & \multirow{3}{*}{Base} & \scriptsize ~~ \bf L31.U10656 & \scriptsize [`:(', `:-)', `:)', `anyway', `solves'] \\
        & & & \scriptsize ~~ \bf L31.U7742 & \scriptsize [`restored', `Accessor', `overwrite', `reuse', `\begin{CJK}{UTF8}{bsmi}：\end{CJK}'] \\
        & & & \scriptsize ~~ \bf L31.U6024 & \scriptsize [`textt', `archivi', `zvuky', `tématu', `lês'] \\
        \cmidrule(lr){3-5}
        & & \multirow{3}{*}{Mmns} & \scriptsize ~~ \bf L28.U2212 & \scriptsize [`etwork', `\foreignlanguage{russian}{окру}', `$\star$', ` ', `Dob'] \\
        & & & \scriptsize ~~ \bf L4.U10613 & \scriptsize [`\foreignlanguage{russian}{Хронологија}', `Archivlink', `$\hookleftarrow$', `$\circ$', `$\blacktriangleright$'] \\
        & & & \scriptsize ~~ \bf L17.U3575 & \scriptsize [`', ` ', `Â', `[...]', `mals'] \\
        \cmidrule(lr){3-5}
        & & \multirow{3}{*}{Ours} & \scriptsize ~~ \bf L29.U7331 & \scriptsize [`Church', `church', `churches', `Kirche', `Kirchen'] \\
        & & & \scriptsize ~~ \bf L27.U7707 & \scriptsize [`Christ', `christ', `Christ', `Christ', `Christians'] \\
        & & & \scriptsize ~~ \bf L21.U1413 & \scriptsize [`church', `\foreignlanguage{russian}{церков}', `churches', `Church', `Religion'] \\
        
        \cmidrule(lr){2-5}
        
        & \multirow{11}{*}{\raisebox{-0.4\height}{mPLUG-Owl2}} & \multirow{3}{*}{Base} & \scriptsize ~~ \bf L31.U1373 & \scriptsize [`', `in', `\textbackslash n', `(', `.'] \\
        & & & \scriptsize ~~ \bf L31.U7491 & \scriptsize [`apparently', `either', `threaten', `towards', `storing'] \\
        & & & \scriptsize ~~ \bf L31.U1563 & \scriptsize [`archivi', `\foreignlanguage{russian}{Kontrola}', `\foreignlanguage{russian}{Хронологија}', `', `'] \\
        \cmidrule(lr){3-5}
        & & \multirow{3}{*}{Mmns} & \scriptsize ~~ \bf L15.U8368 & \scriptsize [`yard', `ill', `go', `mouse', `ments'] \\
        & & & \scriptsize ~~ \bf L19.U1434 & \scriptsize [`snow', `ice', `Snow', `winter', `Winter'] \\
        & & & \scriptsize ~~ \bf L13.U420 & \scriptsize [`church', `Church', `ric', `cho', `uti'] \\
        \cmidrule(lr){3-5}
        & & \multirow{3}{*}{Ours} & \scriptsize ~~ \bf L25.U911 & \scriptsize [`faith', `religion', `relig', `religious', `Relig'] \\
        & & & \scriptsize ~~ \bf L29.U5136 & \scriptsize [`Church', `church', `churches', `\foreignlanguage{russian}{Kirche}', `chiesa'] \\
        & & & \scriptsize ~~ \bf L31.U7266 & \scriptsize [`religious', `Relig', `prayer', `spiritual', `pray'] \\
        \bottomrule
    \end{tabular}
    \caption{An example result shown with top-3 neurons selected by different methods. We report results of the concept \textit{church}. For each neuron, we record its top-5 relative tokens.}
    \label{tab:compared_example}
\end{table*}

\subsection{Identifying Multi-Modal Neurons}
We employ methodology described in \S \ref{subsec:finding_vision_language_skill_neurons} to identify multi-modal neurons in multi-modal LLMs. Figure \ref{fig:distribution_k_50_100_1000} shows the distribution of unique multi-modal neurons. We can see that our multi-modal neurons widely occur in higher layers, which is consistent with previous works~\citep{wang-wen-etal2022skill, dai2022kn}. To further explore characteristics of the multi-modal neurons, we conduct a series of experiments based on them.

\subsection{Are Multi-Modal Neurons Sensitive to Certain Concepts?}
\label{subsec:sensitivity}

We now discuss whether multi-modal neurons are sensitive to certain concepts from four perspectives: (1) Whether multi-modal neurons correspond to \textbf{visual} concepts (\S \ref{subsubsec:tracing_focus}). (2) Whether multi-modal neurons correspond to \textbf{textual} concepts (\S \ref{subsubsec:determining_textual_meanings}). (3) Whether the correspondence between multi-modal neurons and semantic concepts remains constant despite changes in the \textbf{same} image (\S \ref{subsubsec:position_invariance}). (4) Whether the correspondence between multi-modal neurons and semantic concepts remains constant despite changes in \textbf{different} images (\S \ref{subsubsec:cross_images_invariance}).

\subsubsection{Tracing Focus of Neurons in Images}
\label{subsubsec:tracing_focus}

We take the activations of multi-modal neurons at image patch tokens, scale them by bilinear interpolation, and plot the heatmap and binary mask. Implementation details are shown in appendix \ref{appendix_subsec:tracing_feature_regions_in_images}. As the square root of the number of image patch tokens in InstructBLIP and mPLUG-Owl2 is irrational, we only conduct experiments on LLaVA. Table \ref{tab:visual_example} shows an example. We can see that multi-modal neurons mainly focus on image regions that containing corresponding concepts, and pay less attention to other unrelated area. They reliably highlight the semantically pertinent areas throughout.

\begin{table}[!t]
\small
\centering
\renewcommand\arraystretch{1}
\begin{tabular}{ccc@{\hspace{10pt}}c@{\hspace{10pt}}c@{\hspace{10pt}}}
\toprule
\textbf{Model} & \textbf{Method} & \textbf{BS} & \textbf{MS} & \textbf{BRT}\\
\midrule
\multirow{3}{*}{LLaVA} & Base & 0.236 & 0.664 & 0.086 \\
& Mmns & 0.652 & 0.678 & 0.100 \\
& Ours & \bf 0.794 & \bf 0.730 & \bf 0.214 \\
\midrule
\multirow{3}{*}{InstructBLIP} & Base & 0.626 & 0.656 & 0.071 \\
& Mmns & 0.339 & 0.663 & 0.089 \\
& Ours & \bf 0.726 & \bf 0.706 & \bf 0.160 \\
\midrule
\multirow{3}{*}{mPLUG-Owl2} & Base & 0.360 & 0.664 & 0.068 \\
& Mmns & 0.620 & 0.675 & 0.101 \\
& Ours & \bf 0.730 & \bf 0.715 & \bf 0.183 \\
\bottomrule
\end{tabular}
\caption{Results of metrics including BERTScore (BS), MoverScore (MS) and BLEURT (BRT). For each image, we select top-10 multi-modal neurons for each concept, and we record the mean metrics across all concepts. We ultimately calculate means across all images.}
\label{tab:bertscore}
\end{table}

\begin{figure}[!t]
    \centering
    \includegraphics[width=\linewidth]{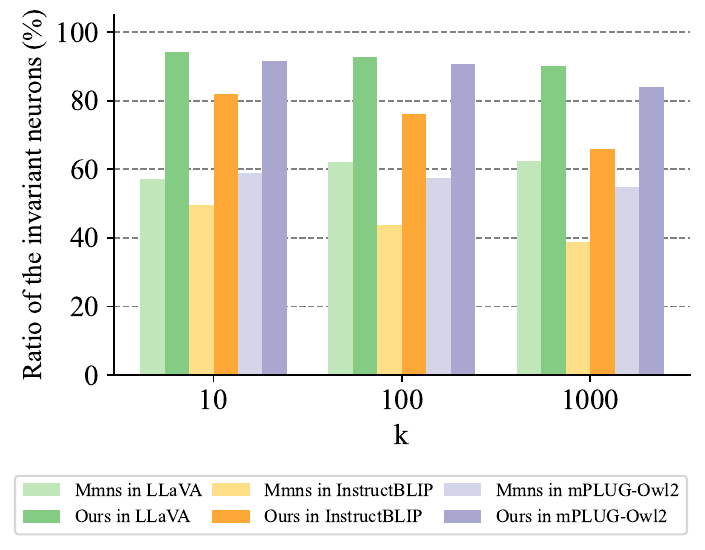} \\
    \caption{Ratios of the invariant neurons in top-$k$ neurons before and after shuffling. For each image, we record the mean ratio across concepts that both exist in original caption and caption generated by shuffled image patches, and then calculate means across all images.}
    \label{fig:invariance}
\end{figure}

\begin{figure}[t]
    \centering
    \includegraphics[width=\linewidth]{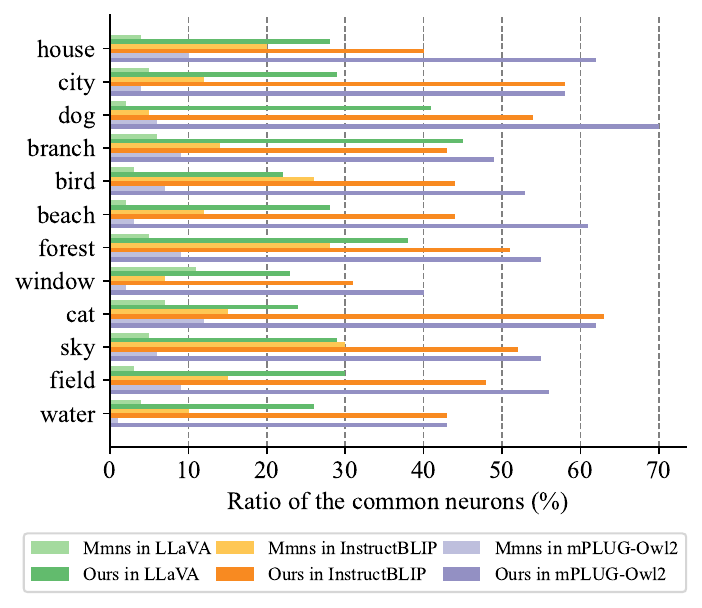}
    \caption{Ratios of the common neurons in top-100 neurons. We set $N=5$ and report results of some concepts that frequently appear in sampled images.}
    \label{fig:commonality_N_5_k_100}
\end{figure}

\begin{figure}[!t]
    \centering
    \begin{varwidth}[t]{\textwidth}
    \vspace*{5pt}
    \includegraphics[width=0.16\linewidth]{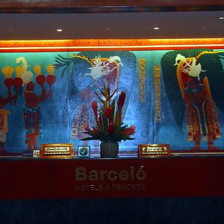} \\[5pt]
    \parbox{2.6cm}{\scriptsize LLaVA: a \textbf{hotel} \textbf{lobby} with a \textbf{reception} \textbf{desk}, a potted \textbf{plant}, and a large \textbf{wall} \textbf{mural} featuring \textbf{angels}.}
    \end{varwidth}
    \hspace{2pt}
    \begin{varwidth}[t]{\textwidth}
    \centering
    \vspace*{0pt}
    \includegraphics[width=0.3\linewidth]{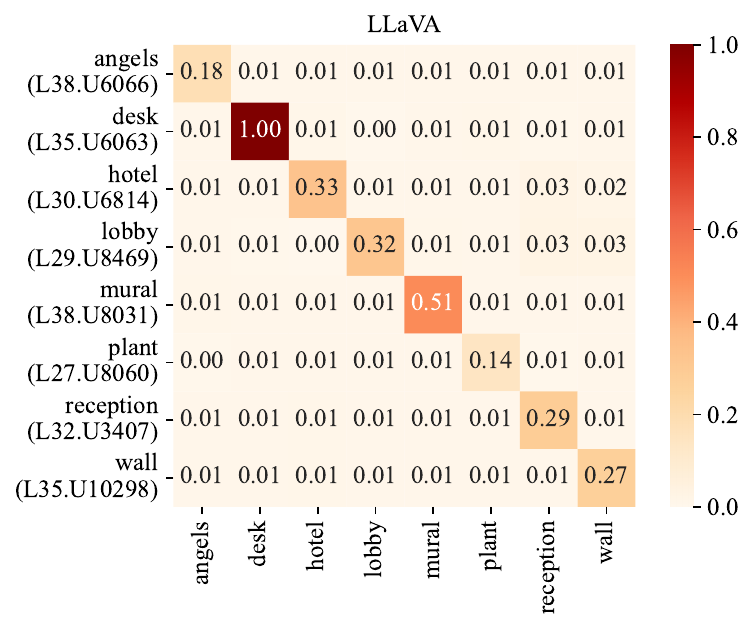}
    \end{varwidth}
    \caption{Heatmap of the scores (after normalization) of multi-modal neurons corresponding to specific concepts when encoding different contents in an example image. The x-axis represents concepts in the given image, and y-axis represents the top-1 neuron corresponding to each concept, respectively. Darker blocks indicate higher scores, which means higher relevance.}
    \label{fig:659_heatmap_vsns}
\end{figure}

\begin{table}[t]
\small
\centering
\renewcommand\arraystretch{1}
\begin{tabular}{cl@{\hspace{6pt}}c@{\hspace{6pt}}c@{\hspace{6pt}}c@{\hspace{6pt}}c@{\hspace{6pt}}}
\toprule
\multicolumn{1}{c}{\textbf{Model}} & \multicolumn{1}{c}{\textbf{Type}} & \textbf{S@1} & \textbf{S@5} & \textbf{S@10} & \textbf{S@50} \\
\midrule
\multirow{2}{*}{LLaVA} & Related & 3.549 & 2.920 & 2.333 & 0.467 \\
& Random & 0.018 & 0.012 & 0.014 & 0.003 \\
\midrule
\multirow{2}{*}{InstructBLIP} & Related & 2.504 & 2.133 & 1.774 & 0.355 \\
& Random & 0.005 & 0.007 & 0.008 & 0.002 \\
\midrule
\multirow{2}{*}{mPLUG-Owl2} & Related & 1.949 & 1.637 & 1.295 & 0.259 \\
& Random & 0.002 & 0.003 & 0.003 & 0.001 \\
\bottomrule
\end{tabular}
\caption{Average scores that multi-modal neurons contribute to related concepts and random concepts. We report average scores with $m = 1, 5, 10, 50$, which are denoted as S@1, S@5, S@10 and S@50, respectively.}
\label{tab:specificity_score}
\end{table}

\begin{table}[!t]
\small
\centering
\renewcommand\arraystretch{0.3}
    \begin{tabular}{c@{\hspace{10pt}}p{5.75cm}}
    \toprule
    \multicolumn{2}{c}{\textbf{Image \& Original output}} \\
    \midrule
    \multicolumn{2}{c}{\includegraphics[width=0.2\linewidth]{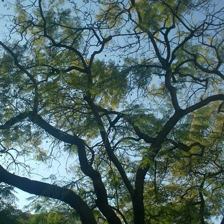} \hspace{5pt}\begin{varwidth}[f]{\textwidth}
        \vspace*{-38pt}\makecell[l]{\scriptsize LLaVA: a \textbf{tree} with many \textbf{branches} and\\[-3pt]\scriptsize \phantom{LLaVA:~}\textbf{leaves}, set against a blue \textbf{sky}.}\end{varwidth}} \\
    \midrule
    \textbf{Concept} & \multicolumn{1}{c}{\textbf{Perturbed model output}} \\
    \midrule
    \multirow{3}{*}{\scriptsize tree} & \scriptsize a Hamon's Garden, featuring a Hamon' the S the Hamon's Garden, featuring a Hamon's the S the Hamon's ... \\
    \midrule
    \scriptsize branches & \scriptsize ameshupelageaameshupelageaamesh... \\
    \midrule
    \multirow{3}{*}{\scriptsize leaves} & \scriptsize a tree with branches spread out, surrounded by tree branches and Homosassa, Florida, and the things around it. \\
    \midrule
    \multirow{5}{*}{\scriptsize sky} & \scriptsize a tree with leaves, possibly a palm tree, with a large and sturdy trunk, surrounded by a large, vibrant, and colorful body of leaves. \\
    \midrule
    {\scriptsize \textit{random}} & \scriptsize a tree with many branches and leaves, set against a blue sky. \\
    \bottomrule
    \end{tabular}
\caption{Perturbation results of LLaVA. For each concept in the image, we only perturb the top-5 multi-modal neurons. For comparison, we report a result of perturbing the same number of random chosen neurons.}
\label{tab:perturb_example}
\end{table}

\subsubsection{Textual Meanings of Neurons}
\label{subsubsec:determining_textual_meanings}
We then verify whether our multi-modal neurons can represent textual meanings. Considering the multiplication of the unembedding matrix and the second layer of FFN is regarded as a projection from the activation of the neurons to probability distributions of the token vocabulary, we empirically sort rows correspond to multi-modal neurons and pick out the top-10 tokens as each neuron represents. We report an example in Table \ref{tab:compared_example}.  We can find that the baseline and Mmns choose the neurons that are hardly correlated with concepts, whereas our method can more precisely identify neurons representing semantic meanings in comparison to them. More examples are shown in appendix \ref{appendix_subsec:finding_vision_language_skill_neurons}.

To provide stronger evidence, we measure metrics of semantic sensitivity mentioned in \S \ref{subsec:evaluation_metrics}. Table \ref{tab:bertscore} shows the mean results. Our method achieve higher scores than Mmns and the baseline, which demonstrates that our selected neurons are more consistent with corresponding concepts.

\subsubsection{Region Invariance of Neurons}
\label{subsubsec:position_invariance}
If multi-modal neurons are exactly sensitive to certain concepts, they shall be invariant when the input sequence of image patches is changed. To quantify the region invariance of the neurons, we calculate the ratio of invariant neurons in top-$k$ neurons when shuffling (see Eq.~\ref{eq:formula10}). The mean results are shown in Figure \ref{fig:invariance}. Our method significantly receives higher ratios of the invariant neurons than Mmns, which indicates our selected multi-modal neurons possess a stronger region invariance.

\begin{table}[t]
\small
    \centering
    \renewcommand\arraystretch{0.2}
    \begin{tabular}{cc@{\hspace{8pt}}p{4.15cm}@{\hspace{4pt}}}
        \toprule
        \multicolumn{3}{c}{\textbf{Image \& Original output}} \\
        \midrule
        \multicolumn{3}{c}{\includegraphics[width=0.2\linewidth]{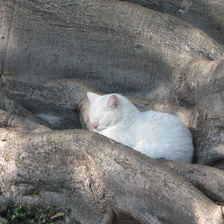} \hspace{5pt}\begin{varwidth}[f]{\textwidth}
        \vspace*{-37pt}\makecell[l]{\parbox{4cm}{\scriptsize  LLaVA: a white \textbf{cat} sleeping in a tree.} \\ \parbox{5cm}{\scriptsize  InstructBLIP: a white \textbf{cat} sleeping in a tree.} \\ \parbox{5.5cm}{\scriptsize  mPLUG-Owl2: a white \textbf{cat} sleeping on a tree branch.}}\end{varwidth}} \\
        \midrule
        \bf Model & \bf Target & \multicolumn{1}{c}{\bf Edited model output} \\
        \midrule
        \scriptsize \multirow{14}{*}{\raisebox{-0.9\height}{LLaVA}} & \scriptsize {monkey} & \scriptsize a white \textbf{monkey} sleeping in a tree. \\
        \cmidrule(lr){2-3}
        & \scriptsize {clock} & \scriptsize a white \textbf{clock} sitting on a tree stump. \\
        \cmidrule(lr){2-3}
        & \scriptsize {iPhone} & \scriptsize a white \textbf{iPhone} lying on a tree stump. \\
        \cmidrule(lr){2-3}
        & \scriptsize {food} & \scriptsize a white \textbf{food} in a tree. \\
        
        \cmidrule(lr){1-3}
        
        \scriptsize \multirow{14}{*}{\raisebox{-0.9\height}{InstructBLIP}} & \scriptsize {monkey} & \scriptsize a white \textbf{monkey} sleeping in a tree. \\
        \cmidrule(lr){2-3}
        & \scriptsize {clock} & \scriptsize a white \textbf{clock} sleeping in a tree. \\
        \cmidrule(lr){2-3}
        & \scriptsize {iPhone} & \scriptsize a white \textbf{iPhone} 3Gs sitting on a tree stump. \\
        \cmidrule(lr){2-3}
        & \scriptsize {food} & \scriptsize a white \textbf{food} sleeping in a tree. \\
        
        \cmidrule(lr){1-3}
        
        \scriptsize \multirow{14}{*}{\raisebox{-0.9\height}{mPLUG-Owl2}} & \scriptsize {monkey} & \scriptsize a white \textbf{monkey} sleeping on a tree branch. \\
        \cmidrule(lr){2-3}
        & \scriptsize {clock} & \scriptsize a \textbf{clock} clocking in a tree trunk. \\
        \cmidrule(lr){2-3}
        & \scriptsize {iPhone} & \scriptsize a white \textbf{iPhone} sitting on a tree branch. \\
        \cmidrule(lr){2-3}
        & \scriptsize {food} & \scriptsize a white \textbf{food} food sleeping on a tree branch. \\
        \bottomrule
        
    \end{tabular}
    \caption{Knowledge editing results of an example. We choose to edit concept \textit{cat} to 4 target concepts. Target concepts are in bold in the edited model output.}
    \label{tab:editing_example}
\end{table}

\begin{table}[t]
    \small
    \centering
    \renewcommand\arraystretch{0.5}
    \begin{tabular}{ccp{4.2cm}}
        \toprule
        \multicolumn{3}{l}{\textbf{Source concept}: bird} \\
        \midrule
        \bf Image & \bf Target & \multicolumn{1}{c}{\bf Edited LLaVA's output} \\
        \midrule
        \multirow{1}{*}{\raisebox{0.0\height}{\makecell{\begin{overpic}[scale=0.16]{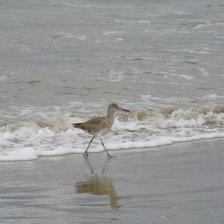} \put(70, 5){\color{red}(a)}\end{overpic}}}} & \scriptsize {None} & \scriptsize a \textbf{bird} walking on the beach near the water. \\
        \cmidrule(lr){2-3}
        & \scriptsize {cat} & \scriptsize a \textbf{cat} walking on the beach near the water. \\
        \cmidrule(lr){2-3}
        & \scriptsize \multirow{2}{*}{\raisebox{-0.3\height}{horse}} & \scriptsize a \textbf{horse} on the beach, walking through the water and enjoying the waves. \\
        \cmidrule(lr){1-3}
        \multirow{1}{*}{\raisebox{-0.8\height}{\makecell{\begin{overpic}[scale=0.16]{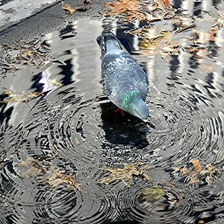} \put(70, 5){\color{red}(b)}\end{overpic}}}} & \scriptsize \multirow{2}{*}{\raisebox{-0.3\height}{None}} & \scriptsize a \textbf{bird}, possibly a pigeon, standing in a puddle of water on a city street. \\
        \cmidrule(lr){2-3}
        & \scriptsize {cat} & \scriptsize a \textbf{cat} sitting in a puddle of water. \\
        \cmidrule(lr){2-3}
        & \scriptsize \raisebox{-0.8\height}{horse} & \scriptsize a \textbf{horse} in a pond, surrounded by leaves and water. \\
        \cmidrule(lr){1-3}
        \multirow{5}{*}{\raisebox{-1.0\height}{\makecell{\begin{overpic}[scale=0.16]{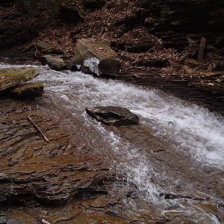} \put(70, 5){\color{red}(c)}\end{overpic}}}} & \scriptsize \multirow{2}{*}{\raisebox{-0.3\height}{None}} & \scriptsize a river flowing through a rocky area, with a waterfall and a rocky cliff. \\
        \cmidrule(lr){2-3}
        & \scriptsize \multirow{2}{*}{\raisebox{-0.3\height}{cat}} & \scriptsize a river flowing through a rocky area, with a waterfall and a rocky cliff. \\
        \cmidrule(lr){2-3}
        & \scriptsize \multirow{2}{*}{\raisebox{-0.3\height}{horse}} & \scriptsize a river flowing through a rocky area, with a waterfall and a rocky cliff. \\
        \bottomrule
    \end{tabular}
    \caption{Edited LLaVA's output of different images. We select \textit{bird} as source concept, choose \textit{cat} and \textit{horse} as target concept (\textit{None} means no editing), and modify model parameters based on image (a). We then test the edited model on another two images, where image (b) contains the source concept \textit{bird} and image (c) doesn't.}
    \label{tab:editing_comparison}
\end{table}

\subsubsection{Cross-Images Invariance of Neurons}
\label{subsubsec:cross_images_invariance}
As for cross-images invariance, same neurons shall occur in different images that carry similar semantic information. To verify cross-images invariance of multi-modal neurons, we calculate the ratio of common neurons by Eq.~\ref{eq:formula11}. The results of Mmns and our method are shown in Figure \ref{fig:commonality_N_5_k_100}. Our multi-modal neurons significantly outperform Mmns. Specifically, our method achieves common neuron ratios over 20\% in LLaVA and mostly over 40\% in InstructBLIP and mPLUG-Owl2, which is substantially higher than Mmns that attain ratios mainly under 10\% in LLaVA, under 30\% in InstructBLIP and under 20\% in mPLUG-Owl2. We report more results with different $N$ and $k$ in appendix \ref{subsec:commonality_of_vision_language_skill_neurons}.

\subsection{Are Multi-Modal Neurons Specific?}
\label{subsec:specificity}
For multi-modal neurons, claiming indiscriminate sensitivity to all concepts does not sufficiently demonstrate their functional role within the model. As such, we investigate their specificity. We record the scores of multi-modal neurons that correspond to their specific textual meanings when encoding other different concepts in the same image. Figure \ref{fig:659_heatmap_vsns} shows an example. Additional examples are provided in appendix \ref{subsec:specificity_of_vision_language_skill_neurons}. We can see that when encoding a specific concept, the top-1 multi-modal neuron receives a higher score than irrelevant concepts. We also adopt a metric to quantify the specificity of neurons (see \S \ref{subsec:evaluation_metrics}). The results are shown in Table \ref{tab:specificity_score}, from which we can find that neurons significantly get higher scores to those related concepts than to unrelated concepts, proving their specificity.

\subsection{Do Multi-Modal Neurons Causally Affect Output?}
\label{subsec:causal_effect}

\noindent\textbf{Perturbation Study:}  Previous works~\citep{mitchell2022fast, meng2022locating, meng2022mass} have shown that applying directional editing to FFNs significantly change the model output. Inspired by these, we try to perturb multi-modal neurons. Specifically, for each concept in each image, we add a Gaussian noise $(\mu=0 \text{~and~} \sigma=0.5)$ to the $i$-th row of the second layer of FFN at layer $l$. Table \ref{tab:perturb_example} shows an example when perturbing neurons in LLaVA. We can see that perturbing multi-modal neurons really makes a difference in model output, while simply perturbing few random neurons has no impact. Furthermore, we note that applying perturbation on neurons sometimes makes the corresponding token disappear in output and provides some new tokens, while sometimes results in meaningless output (e.g., in Table \ref{tab:perturb_example}, when we perturb concepts `leaves' and `sky', the model can generate fluent output without `leaves' and `sky', but it is confused when we perturb concepts `tree' and `branches'). The former phenomenon piques our curiosity regarding the potential possibility that a well-designed alteration may substitute for Gaussian noise to enable knowledge editing of model output.

\noindent\textbf{Knowledge Editing:}  We hypothesize that replacing the Gaussian noise with an elaborate alteration can achieve a knowledge editing. Accordingly, we design an efficient algorithm (see Algorithm \ref{algo:targeted_modification}) that edits weights of the second layer of FFNs. Table \ref{tab:editing_example} shows an example, where we guide the model to generate a different concept from the original concept. We find that model drops the source concept and successfully generates the target concept, which did not appear in original output. To prove effectiveness of our method, we evaluate the edited model on other different images, as shown in Table \ref{tab:editing_comparison}. We find that when we input another image that contains the same source concept, the edited model will identify it and generate the target concept, while an unrelated image will not be affected.

\section{Related Work}
\label{sec:rw}
\noindent\textbf{Identifying Neurons in Deep Neural Networks:}\quad There has been growing interest in interpreting and analyzing the inner workings of deep neural networks. Prior works have sought to characterize what types of information are encoded in individual neurons. \citet{koh2020concept} proposes a technique for identifying ``concept neurons'' that detect semantic concepts in vision models. \citet{dai2022kn} discusses the discovery of ``knowledge neurons'' which encode specific commonsense knowledge automatically learned during pre-training, while \citet{wang-wen-etal2022skill} proposes a method to identify ``skill neurons'' in pre-trained Transformer-based language models that are heavily involved in specific tasks. Recently,~\citet{schwettmann2023multimodal} introduces a procedure for identifying ``multimodal neurons'', which explain how LLMs convert visual representations into corresponding texts.

\noindent\textbf{Analysing Pre-Trained Transformers:}\quad Over the past decade, we have witnessed the fast development and vast success of deep neural network architectures in many communities~\cite{yang2024learning,di2024hyper,yang2022video,yang2021deconfounded,yang2018person,yang2024robust,yang2020tree,song2024emotional}. Transformer~\citep{vaswani2017attention} is one of the most successful architectures and Transformer-based models have attracted a large amount of studies~\citep{li2023redundancy,li2023transformer}. Prior works have focused on the function and mechanism of self-attention modules~\citep{voitaetal2019analyzing, clark2019what, hao2021self}, while some works emphasize the significance of feed-forward layers in Transformer~\citep{pressetal2020improving, geva2020transformer, dai2022kn}. Among these, some works probe Transformer representations to quantify their encoding of linguistic information~\citep{peters2018dissecting, niven2019probing, yun2019transformers}.

\section{Conclusion}
\label{sec:conc}
We propose a new method to identify multi-modal neurons in Transformer-based multi-modal LLMs. We also introduce a knowledge editing approach based on the identified neurons, which achieves a knowledge editing from a specific token to another designative token. We highlight three critical properties of multi-modal neurons by four well-designed quantitative evaluation metrics through extensive experiments. Both quantitative and qualitative experiments validate the explanatory powers of our multi-modal neurons. This work provides illuminating perspectives on multi-modal LLMs and stimulates additional explanatory artificial intelligence studies emphasizing model interpretability.

\section*{Limitations}
While this work provides new insights into interpreting multi-modal large language models, there are several limitations that should be acknowledged: (1) We only conduct experiments on LLaVA, InstructBLIP and mPLUG-Owl2, while other Transformer-based models may also be possible to be explained by our multi-modal neurons. Besides the Transformer architecture, it is still unclear whether neurons exist in other multi-modal large language models based on different architectures and requires further explorations. (2) We only focus on neurons in feed-forward networks in Transformer and omit other parts like the neurons in self-attention heads, which may also contribute to identify image features and generate output. (3) When analysing multi-modal neurons, we only consider the role of a single neuron. We expect future works can explore how multiple neurons jointly influence the model. (4) As our multi-modal knowledge editing method is based on changing the probability distribution of the generated token, we only achieve a transformation from a single source token to another single designative token, which is still insufficient, since there are a large amount of words consist of multiple tokens. We will investigate editing multiple tokens in our future work.

Further addressing these limitations through broader and more methodologically rigorous studies would help advance knowledge in interpretability of multi-modal large language models.

\section*{Acknowledgements}
This work was supported by the National Natural Science Foundation of China (NSFC) under Grant U22A2094 and Grant 62272435, and also supported by the advanced computing resources provided by the Supercomputing Center of the USTC. We also acknowledge the support of GPU cluster built by MCC Lab of Information Science and Technology Institution, USTC.

\bibliography{custom}

\begin{thebibliography}{49}
\expandafter\ifx\csname natexlab\endcsname\relax\def\natexlab#1{#1}\fi

\bibitem[{Bau et~al.(2017)Bau, Zhou, Khosla, Oliva, and Torralba}]{netdissect2017}
David Bau, Bolei Zhou, Aditya Khosla, Aude Oliva, and Antonio Torralba. 2017.
\newblock \href {https://openaccess.thecvf.com/content_cvpr_2017/html/Bau_Network_Dissection_Quantifying_CVPR_2017_paper.html} {Network dissection: Quantifying interpretability of deep visual representations}.
\newblock In \emph{Proceedings of the IEEE Conference on Computer Vision and Pattern Recognition (CVPR)}.

\bibitem[{Chiang et~al.(2023)Chiang, Li, Lin, Sheng, Wu, Zhang, Zheng, Zhuang, Zhuang, Gonzalez, Stoica, and Xing}]{vicuna2023}
Wei-Lin Chiang, Zhuohan Li, Zi~Lin, Ying Sheng, Zhanghao Wu, Hao Zhang, Lianmin Zheng, Siyuan Zhuang, Yonghao Zhuang, Joseph~E. Gonzalez, Ion Stoica, and Eric~P. Xing. 2023.
\newblock \href {https://lmsys.org/blog/2023-03-30-vicuna/} {Vicuna: An open-source chatbot impressing gpt-4 with 90\%* chatgpt quality}.

\bibitem[{Clark et~al.(2019)Clark, Khandelwal, Levy, and Manning}]{clark2019what}
Kevin Clark, Urvashi Khandelwal, Omer Levy, and Christopher~D. Manning. 2019.
\newblock \href {https://doi.org/10.18653/v1/W19-4828} {What does {BERT} look at? an analysis of {BERT}{'}s attention}.
\newblock In \emph{Proceedings of the 2019 ACL Workshop BlackboxNLP: Analyzing and Interpreting Neural Networks for NLP}, pages 276--286, Florence, Italy. Association for Computational Linguistics.

\bibitem[{Dai et~al.(2022)Dai, Dong, Hao, Sui, Chang, and Wei}]{dai2022kn}
Damai Dai, Li~Dong, Yaru Hao, Zhifang Sui, Baobao Chang, and Furu Wei. 2022.
\newblock \href {https://doi.org/10.18653/v1/2022.acl-long.581} {Knowledge neurons in pretrained transformers}.
\newblock In \emph{Proceedings of the 60th Annual Meeting of the Association for Computational Linguistics (Volume 1: Long Papers)}, pages 8493--8502, Dublin, Ireland. Association for Computational Linguistics.

\bibitem[{Dai et~al.(2023)Dai, Li, Li, Tiong, Zhao, Wang, Li, Fung, and Hoi}]{instructblip}
Wenliang Dai, Junnan Li, Dongxu Li, Anthony Meng~Huat Tiong, Junqi Zhao, Weisheng Wang, Boyang Li, Pascale Fung, and Steven Hoi. 2023.
\newblock \href {http://arxiv.org/abs/2305.06500} {Instructblip: Towards general-purpose vision-language models with instruction tuning}.

\bibitem[{Di et~al.(2024)Di, Yang, Luo, Xue, Chen, Yang, and Gao}]{di2024hyper}
Donglin Di, Jiahui Yang, Chaofan Luo, Zhou Xue, Wei Chen, Xun Yang, and Yue Gao. 2024.
\newblock \href {https://arxiv.org/abs/2403.09236} {Hyper-3dg: Text-to-3d gaussian generation via hypergraph}.
\newblock \emph{arXiv preprint arXiv:2403.09236}.

\bibitem[{Duan et~al.(2023)Duan, Cheng, Wang, Wang, Zavalny, Xu, Kailkhura, and Xu}]{duan2023shifting}
Jinhao Duan, Hao Cheng, Shiqi Wang, Chenan Wang, Alex Zavalny, Renjing Xu, Bhavya Kailkhura, and Kaidi Xu. 2023.
\newblock \href {https://arxiv.org/abs/2307.01379v2} {Shifting attention to relevance: Towards the uncertainty estimation of large language models}.
\newblock \emph{arXiv preprint arXiv:2307.01379}.

\bibitem[{Fang et~al.(2023)Fang, Wang, Xie, Sun, Wu, Wang, Huang, Wang, and Cao}]{Fang_2023_CVPR}
Yuxin Fang, Wen Wang, Binhui Xie, Quan Sun, Ledell Wu, Xinggang Wang, Tiejun Huang, Xinlong Wang, and Yue Cao. 2023.
\newblock \href {https://openaccess.thecvf.com/content/CVPR2023/html/Fang_EVA_Exploring_the_Limits_of_Masked_Visual_Representation_Learning_at_CVPR_2023_paper.html} {Eva: Exploring the limits of masked visual representation learning at scale}.
\newblock In \emph{Proceedings of the IEEE/CVF Conference on Computer Vision and Pattern Recognition (CVPR)}, pages 19358--19369.

\bibitem[{Geng et~al.(2023)Geng, Gudibande, Liu, Wallace, Abbeel, Levine, and Song}]{koala_blogpost_2023}
Xinyang Geng, Arnav Gudibande, Hao Liu, Eric Wallace, Pieter Abbeel, Sergey Levine, and Dawn Song. 2023.
\newblock \href {https://bair.berkeley.edu/blog/2023/04/03/koala/} {Koala: A dialogue model for academic research}.
\newblock Blog post.

\bibitem[{Geva et~al.(2021)Geva, Schuster, Berant, and Levy}]{geva2020transformer}
Mor Geva, Roei Schuster, Jonathan Berant, and Omer Levy. 2021.
\newblock \href {https://doi.org/10.18653/v1/2021.emnlp-main.446} {Transformer feed-forward layers are key-value memories}.
\newblock In \emph{Proceedings of the 2021 Conference on Empirical Methods in Natural Language Processing}, pages 5484--5495, Online and Punta Cana, Dominican Republic. Association for Computational Linguistics.

\bibitem[{Hao et~al.(2021)Hao, Dong, Wei, and Xu}]{hao2021self}
Yaru Hao, Li~Dong, Furu Wei, and Ke~Xu. 2021.
\newblock \href {https://ojs.aaai.org/index.php/AAAI/article/view/17533} {Self-attention attribution: Interpreting information interactions inside transformer}.
\newblock In \emph{Proceedings of the AAAI Conference on Artificial Intelligence}, volume~35, pages 12963--12971.

\bibitem[{Huang et~al.(2023)Huang, Song, Wang, Chen, and Ma}]{huang2023look}
Yuheng Huang, Jiayang Song, Zhijie Wang, Huaming Chen, and Lei Ma. 2023.
\newblock \href {https://arxiv.org/abs/2307.10236v3} {Look before you leap: An exploratory study of uncertainty measurement for large language models}.
\newblock \emph{arXiv preprint arXiv:2307.10236}.

\bibitem[{Koh et~al.(2020)Koh, Nguyen, Tang, Mussmann, Pierson, Kim, and Liang}]{koh2020concept}
Pang~Wei Koh, Thao Nguyen, Yew~Siang Tang, Stephen Mussmann, Emma Pierson, Been Kim, and Percy Liang. 2020.
\newblock \href {https://proceedings.mlr.press/v119/koh20a.html} {Concept bottleneck models}.
\newblock In \emph{Proceedings of the 37th International Conference on Machine Learning}, volume 119 of \emph{Proceedings of Machine Learning Research}, pages 5338--5348. PMLR.

\bibitem[{Li et~al.(2023{\natexlab{a}})Li, Li, Savarese, and Hoi}]{li2023blip2}
Junnan Li, Dongxu Li, Silvio Savarese, and Steven Hoi. 2023{\natexlab{a}}.
\newblock \href {https://icml.cc/virtual/2023/poster/25182} {{BLIP-2:} bootstrapping language-image pre-training with frozen image encoders and large language models}.
\newblock In \emph{ICML}.

\bibitem[{Li et~al.(2023{\natexlab{b}})Li, Li, Guo, Yang, and Wang}]{li2023transformer}
Kun Li, Jiaxiu Li, Dan Guo, Xun Yang, and Meng Wang. 2023{\natexlab{b}}.
\newblock \href {https://dl.acm.org/doi/abs/10.1145/3587251} {Transformer-based visual grounding with cross-modality interaction}.
\newblock \emph{ACM Transactions on Multimedia Computing, Communications and Applications}, 19(6):1--19.

\bibitem[{Li et~al.(2023{\natexlab{c}})Li, Yang, Zhang, Feng, Wang, and Chua}]{li2023redundancy}
Yicong Li, Xun Yang, An~Zhang, Chun Feng, Xiang Wang, and Tat-Seng Chua. 2023{\natexlab{c}}.
\newblock \href {https://dl.acm.org/doi/abs/10.1145/3581783.3612577} {Redundancy-aware transformer for video question answering}.
\newblock In \emph{Proceedings of the 31st ACM International Conference on Multimedia}, pages 3172--3180.

\bibitem[{Liu et~al.(2023)Liu, Li, Wu, and Lee}]{liu2023visual}
Haotian Liu, Chunyuan Li, Qingyang Wu, and Yong~Jae Lee. 2023.
\newblock \href {https://arxiv.org/abs/2304.08485} {Visual instruction tuning}.
\newblock \emph{arXiv preprint arXiv:2304.08485}.

\bibitem[{Manning et~al.(2014)Manning, Surdeanu, Bauer, Finkel, Bethard, and McClosky}]{manning2014stanford}
Christopher Manning, Mihai Surdeanu, John Bauer, Jenny Finkel, Steven Bethard, and David McClosky. 2014.
\newblock \href {https://doi.org/10.3115/v1/P14-5010} {The {S}tanford {C}ore{NLP} natural language processing toolkit}.
\newblock In \emph{Proceedings of 52nd Annual Meeting of the Association for Computational Linguistics: System Demonstrations}, pages 55--60, Baltimore, Maryland. Association for Computational Linguistics.

\bibitem[{Meng et~al.(2022)Meng, Bau, Andonian, and Belinkov}]{meng2022locating}
Kevin Meng, David Bau, Alex Andonian, and Yonatan Belinkov. 2022.
\newblock \href {https://proceedings.neurips.cc/paper_files/paper/2022/hash/6f1d43d5a82a37e89b0665b33bf3a182-Abstract-Conference.html} {Locating and editing factual associations in gpt}.
\newblock \emph{Advances in Neural Information Processing Systems}, 35:17359--17372.

\bibitem[{Meng et~al.(2023)Meng, Sharma, Andonian, Belinkov, and Bau}]{meng2022mass}
Kevin Meng, Arnab~Sen Sharma, Alex~J Andonian, Yonatan Belinkov, and David Bau. 2023.
\newblock \href {https://openreview.net/forum?id=MkbcAHIYgyS} {Mass-editing memory in a transformer}.
\newblock In \emph{The Eleventh International Conference on Learning Representations}.

\bibitem[{Merullo et~al.(2023)Merullo, Eickhoff, and Pavlick}]{merullo2023language}
Jack Merullo, Carsten Eickhoff, and Ellie Pavlick. 2023.
\newblock \href {http://arxiv.org/abs/2305.16130} {Language models implement simple word2vec-style vector arithmetic}.

\bibitem[{Mitchell et~al.(2022)Mitchell, Lin, Bosselut, Finn, and Manning}]{mitchell2022fast}
Eric Mitchell, Charles Lin, Antoine Bosselut, Chelsea Finn, and Christopher~D Manning. 2022.
\newblock \href {https://openreview.net/pdf?id=0DcZxeWfOPt} {Fast model editing at scale}.
\newblock In \emph{International Conference on Learning Representations}.

\bibitem[{Niven and Kao(2019)}]{niven2019probing}
Timothy Niven and Hung-Yu Kao. 2019.
\newblock \href {https://www.aclweb.org/anthology/P19-1459} {Probing neural network comprehension of natural language arguments}.
\newblock In \emph{Proceedings of the 57th Conference of the Association for Computational Linguistics}, pages 4658--4664, Florence, Italy. Association for Computational Linguistics.

\bibitem[{Olsson et~al.(2022)Olsson, Elhage, Nanda, Joseph, DasSarma, Henighan, Mann, Askell, Bai, Chen et~al.}]{olsson2022context}
Catherine Olsson, Nelson Elhage, Neel Nanda, Nicholas Joseph, Nova DasSarma, Tom Henighan, Ben Mann, Amanda Askell, Yuntao Bai, Anna Chen, et~al. 2022.
\newblock \href {https://arxiv.org/abs/2209.11895} {In-context learning and induction heads}.
\newblock \emph{arXiv preprint arXiv:2209.11895}.

\bibitem[{Ordonez et~al.(2011)Ordonez, Kulkarni, and Berg}]{Ordonez:2011:im2text}
Vicente Ordonez, Girish Kulkarni, and Tamara~L. Berg. 2011.
\newblock \href {https://proceedings.neurips.cc/paper_files/paper/2011/hash/5dd9db5e033da9c6fb5ba83c7a7ebea9-Abstract.html} {Im2text: Describing images using 1 million captioned photographs}.
\newblock In \emph{Neural Information Processing Systems ({NIPS})}.

\bibitem[{Peters et~al.(2018)Peters, Neumann, Zettlemoyer, and Yih}]{peters2018dissecting}
Matthew~E Peters, Mark Neumann, Luke Zettlemoyer, and Wen-tau Yih. 2018.
\newblock \href {https://arxiv.org/abs/1808.08949v2} {Dissecting contextual word embeddings: Architecture and representation}.
\newblock \emph{arXiv preprint arXiv:1808.08949}.

\bibitem[{Press et~al.(2020)Press, Smith, and Levy}]{pressetal2020improving}
Ofir Press, Noah~A. Smith, and Omer Levy. 2020.
\newblock \href {https://doi.org/10.18653/v1/2020.acl-main.270} {Improving transformer models by reordering their sublayers}.
\newblock In \emph{Proceedings of the 58th Annual Meeting of the Association for Computational Linguistics}, pages 2996--3005, Online. Association for Computational Linguistics.

\bibitem[{Radford et~al.(2021)Radford, Kim, Hallacy, Ramesh, Goh, Agarwal, Sastry, Askell, Mishkin, Clark, Krueger, and Sutskever}]{radford2021learning}
Alec Radford, Jong~Wook Kim, Chris Hallacy, Aditya Ramesh, Gabriel Goh, Sandhini Agarwal, Girish Sastry, Amanda Askell, Pamela Mishkin, Jack Clark, Gretchen Krueger, and Ilya Sutskever. 2021.
\newblock \href {https://proceedings.mlr.press/v139/radford21a.html} {Learning transferable visual models from natural language supervision}.
\newblock In \emph{Proceedings of the 38th International Conference on Machine Learning}, volume 139 of \emph{Proceedings of Machine Learning Research}, pages 8748--8763. PMLR.

\bibitem[{Robbins and Monro(1951)}]{robbins1951stochastic}
Herbert Robbins and Sutton Monro. 1951.
\newblock A stochastic approximation method.
\newblock \emph{The annals of mathematical statistics}, pages 400--407.

\bibitem[{Schwettmann et~al.(2023)Schwettmann, Chowdhury, Klein, Bau, and Torralba}]{schwettmann2023multimodal}
Sarah Schwettmann, Neil Chowdhury, Samuel Klein, David Bau, and Antonio Torralba. 2023.
\newblock \href {https://openaccess.thecvf.com/content/ICCV2023W/CLVL/html/Schwettmann_Multimodal_Neurons_in_Pretrained_Text-Only_Transformers_ICCVW_2023_paper.html} {Multimodal neurons in pretrained text-only transformers}.
\newblock In \emph{Proceedings of the IEEE/CVF International Conference on Computer Vision}, pages 2862--2867.

\bibitem[{Sellam et~al.(2020)Sellam, Das, and Parikh}]{sellam2020bleurt}
Thibault Sellam, Dipanjan Das, and Ankur Parikh. 2020.
\newblock \href {https://doi.org/10.18653/v1/2020.acl-main.704} {{BLEURT}: Learning robust metrics for text generation}.
\newblock In \emph{Proceedings of the 58th Annual Meeting of the Association for Computational Linguistics}, pages 7881--7892, Online. Association for Computational Linguistics.

\bibitem[{Song et~al.(2024)Song, Guo, Yang, Tang, and Wang}]{song2024emotional}
Peipei Song, Dan Guo, Xun Yang, Shengeng Tang, and Meng Wang. 2024.
\newblock \href {https://ieeexplore.ieee.org/abstract/document/10418849/} {Emotional video captioning with vision-based emotion interpretation network}.
\newblock \emph{IEEE Transactions on Image Processing}.

\bibitem[{Taori et~al.(2023)Taori, Gulrajani, Zhang, Dubois, Li, Guestrin, Liang, and Hashimoto}]{alpaca}
Rohan Taori, Ishaan Gulrajani, Tianyi Zhang, Yann Dubois, Xuechen Li, Carlos Guestrin, Percy Liang, and Tatsunori~B. Hashimoto. 2023.
\newblock Stanford alpaca: An instruction-following llama model.
\newblock \url{https://github.com/tatsu-lab/stanford_alpaca}.

\bibitem[{Touvron et~al.(2023{\natexlab{a}})Touvron, Lavril, Izacard, Martinet, Lachaux, Lacroix, Rozi{\`e}re, Goyal, Hambro, Azhar et~al.}]{touvron2023llama}
Hugo Touvron, Thibaut Lavril, Gautier Izacard, Xavier Martinet, Marie-Anne Lachaux, Timoth{\'e}e Lacroix, Baptiste Rozi{\`e}re, Naman Goyal, Eric Hambro, Faisal Azhar, et~al. 2023{\natexlab{a}}.
\newblock \href {https://arxiv.org/abs/2302.13971} {Llama: Open and efficient foundation language models}.
\newblock \emph{arXiv preprint arXiv:2302.13971}.

\bibitem[{Touvron et~al.(2023{\natexlab{b}})Touvron, Martin, Stone, Albert, Almahairi, Babaei, Bashlykov, Batra, Bhargava, Bhosale et~al.}]{touvron2023llama2}
Hugo Touvron, Louis Martin, Kevin Stone, Peter Albert, Amjad Almahairi, Yasmine Babaei, Nikolay Bashlykov, Soumya Batra, Prajjwal Bhargava, Shruti Bhosale, et~al. 2023{\natexlab{b}}.
\newblock \href {https://arxiv.org/abs/2307.09288v2} {Llama 2: Open foundation and fine-tuned chat models}.
\newblock \emph{arXiv preprint arXiv:2307.09288}.

\bibitem[{Vaswani et~al.(2017)Vaswani, Shazeer, Parmar, Uszkoreit, Jones, Gomez, Kaiser, and Polosukhin}]{vaswani2017attention}
Ashish Vaswani, Noam Shazeer, Niki Parmar, Jakob Uszkoreit, Llion Jones, Aidan~N Gomez, {\L}ukasz Kaiser, and Illia Polosukhin. 2017.
\newblock \href {https://proceedings.neurips.cc/paper_files/paper/2017/hash/3f5ee243547dee91fbd053c1c4a845aa-Abstract.html} {Attention is all you need}.
\newblock \emph{Advances in neural information processing systems}, 30.

\bibitem[{Voita et~al.(2019)Voita, Talbot, Moiseev, Sennrich, and Titov}]{voitaetal2019analyzing}
Elena Voita, David Talbot, Fedor Moiseev, Rico Sennrich, and Ivan Titov. 2019.
\newblock \href {https://doi.org/10.18653/v1/P19-1580} {Analyzing multi-head self-attention: Specialized heads do the heavy lifting, the rest can be pruned}.
\newblock In \emph{Proceedings of the 57th Annual Meeting of the Association for Computational Linguistics}, pages 5797--5808, Florence, Italy. Association for Computational Linguistics.

\bibitem[{Wang et~al.(2022)Wang, Wen, Zhang, Hou, Liu, and Li}]{wang-wen-etal2022skill}
Xiaozhi Wang, Kaiyue Wen, Zhengyan Zhang, Lei Hou, Zhiyuan Liu, and Juanzi Li. 2022.
\newblock \href {https://doi.org/10.18653/v1/2022.emnlp-main.765} {Finding skill neurons in pre-trained transformer-based language models}.
\newblock In \emph{Proceedings of the 2022 Conference on Empirical Methods in Natural Language Processing}, pages 11132--11152, Abu Dhabi, United Arab Emirates. Association for Computational Linguistics.

\bibitem[{Yang et~al.(2024{\natexlab{a}})Yang, Chang, Zhang, Wang, Hong, and Wang}]{yang2024learning}
Xun Yang, Tianyu Chang, Tianzhu Zhang, Shanshan Wang, Richang Hong, and Meng Wang. 2024{\natexlab{a}}.
\newblock \href {https://link.springer.com/article/10.1007/s11263-024-02106-7} {Learning hierarchical visual transformation for domain generalizable visual matching and recognition}.
\newblock \emph{International Journal of Computer Vision}, pages 1--27.

\bibitem[{Yang et~al.(2020)Yang, Dong, Cao, Wang, Wang, and Chua}]{yang2020tree}
Xun Yang, Jianfeng Dong, Yixin Cao, Xun Wang, Meng Wang, and Tat-Seng Chua. 2020.
\newblock \href {https://dl.acm.org/doi/abs/10.1145/3397271.3401151} {Tree-augmented cross-modal encoding for complex-query video retrieval}.
\newblock In \emph{Proceedings of the 43rd international ACM SIGIR conference on research and development in information retrieval}, pages 1339--1348.

\bibitem[{Yang et~al.(2021)Yang, Feng, Ji, Wang, and Chua}]{yang2021deconfounded}
Xun Yang, Fuli Feng, Wei Ji, Meng Wang, and Tat-Seng Chua. 2021.
\newblock \href {https://dl.acm.org/doi/abs/10.1145/3404835.3462823} {Deconfounded video moment retrieval with causal intervention}.
\newblock In \emph{Proceedings of the 44th International ACM SIGIR Conference on Research and Development in Information Retrieval}, pages 1--10.

\bibitem[{Yang et~al.(2022)Yang, Wang, Dong, Dong, Wang, and Chua}]{yang2022video}
Xun Yang, Shanshan Wang, Jian Dong, Jianfeng Dong, Meng Wang, and Tat-Seng Chua. 2022.
\newblock \href {https://ieeexplore.ieee.org/abstract/document/9677948} {Video moment retrieval with cross-modal neural architecture search}.
\newblock \emph{IEEE Transactions on Image Processing}, 31:1204--1216.

\bibitem[{Yang et~al.(2024{\natexlab{b}})Yang, Zeng, Guo, Wang, Dong, and Wang}]{yang2024robust}
Xun Yang, Jianming Zeng, Dan Guo, Shanshan Wang, Jianfeng Dong, and Meng Wang. 2024{\natexlab{b}}.
\newblock Robust video question answering via contrastive cross-modality representation learning.
\newblock \emph{SCIENCE CHINA Information Sciences}.

\bibitem[{Yang et~al.(2018)Yang, Zhou, and Wang}]{yang2018person}
Xun Yang, Peicheng Zhou, and Meng Wang. 2018.
\newblock \href {https://ieeexplore.ieee.org/abstract/document/8445716/} {Person reidentification via structural deep metric learning}.
\newblock \emph{IEEE transactions on neural networks and learning systems}, 30(10):2987--2998.

\bibitem[{Ye et~al.(2023{\natexlab{a}})Ye, Xu, Xu, Ye, Yan, Zhou, Wang, Hu, Shi, Shi et~al.}]{ye2023mplug}
Qinghao Ye, Haiyang Xu, Guohai Xu, Jiabo Ye, Ming Yan, Yiyang Zhou, Junyang Wang, Anwen Hu, Pengcheng Shi, Yaya Shi, et~al. 2023{\natexlab{a}}.
\newblock \href {https://arxiv.org/abs/2304.14178} {mplug-owl: Modularization empowers large language models with multimodality}.
\newblock \emph{arXiv preprint arXiv:2304.14178}.

\bibitem[{Ye et~al.(2023{\natexlab{b}})Ye, Xu, Ye, Yan, Liu, Qian, Zhang, Huang, and Zhou}]{ye2023mplugowl2}
Qinghao Ye, Haiyang Xu, Jiabo Ye, Ming Yan, Haowei Liu, Qi~Qian, Ji~Zhang, Fei Huang, and Jingren Zhou. 2023{\natexlab{b}}.
\newblock \href {https://arxiv.org/abs/2311.04257v2} {mplug-owl2: Revolutionizing multi-modal large language model with modality collaboration}.
\newblock \emph{arXiv preprint arXiv:2311.04257}.

\bibitem[{Yun et~al.(2019)Yun, Bhojanapalli, Rawat, Reddi, and Kumar}]{yun2019transformers}
Chulhee Yun, Srinadh Bhojanapalli, Ankit~Singh Rawat, Sashank~J Reddi, and Sanjiv Kumar. 2019.
\newblock \href {https://arxiv.org/abs/1912.10077v2} {Are transformers universal approximators of sequence-to-sequence functions?}
\newblock \emph{arXiv preprint arXiv:1912.10077}.

\bibitem[{Zhang et~al.(2020)Zhang, Kishore, Wu, Weinberger, and Artzi}]{zhang2019bertscore}
Tianyi Zhang, Varsha Kishore, Felix Wu, Kilian~Q Weinberger, and Yoav Artzi. 2020.
\newblock \href {https://openreview.net/forum?id=SkeHuCVFDr} {Bertscore: Evaluating text generation with bert}.
\newblock In \emph{International Conference on Learning Representations}.

\bibitem[{Zhao et~al.(2019)Zhao, Peyrard, Liu, Gao, Meyer, and Eger}]{zhao2019moverscore}
Wei Zhao, Maxime Peyrard, Fei Liu, Yang Gao, Christian~M. Meyer, and Steffen Eger. 2019.
\newblock \href {https://doi.org/10.18653/v1/D19-1053} {{M}over{S}core: Text generation evaluating with contextualized embeddings and earth mover distance}.
\newblock In \emph{Proceedings of the 2019 Conference on Empirical Methods in Natural Language Processing and the 9th International Joint Conference on Natural Language Processing (EMNLP-IJCNLP)}, pages 563--578, Hong Kong, China. Association for Computational Linguistics.

\end{thebibliography}
\bibliographystyle{acl_natbib}

\appendix

\section{Supplementary Explanation}
\label{sec:supplementary_explanation}
In \S~\ref{subsec:finding_vision_language_skill_neurons}, we illustrate how to identify multi-modal neurons in Transformer-based~\cite{vaswani2017attention} LLMs. We now provide some additional details here.

In Eq.~\ref{eq:formula9}, we use matrix $\mathbf{Q}^l$ to define the contribution score. From the dimensional perspective of $\mathbf{Q}^l$, since $\mathbf{Q}^l \in \mathbb{R}^{d_m\times v}$, where $d_m$ is intermediate size and $v$ is vocab size, each element in $\mathbf{Q}^l$ can be regarded as a contribution of each neuron at layer $l$ to each token in the vocabulary. For instance, the contribution of the $i$-th neuron $u_i$ at layer $l$ to token $t$ is derived from the $i$-th row and $t$-th column of $\mathbf{Q}^l$ (i.e. $\mathbf{Q}^l(i, t)$). From the perspective of the meaning of $\mathbf{Q}^l$, $\mathbf{Q}^l$ is consistent with the probability distribution when predicting, where we prove it through Eq.~\ref{eq:formula7} and Eq.~\ref{eq:formula8}.

In Eq.~\ref{eq:formula7}, we disassemble the generation procedure of the LLM. We first decompose the hidden states at the last layer $\mathbf{h}^L_{-1}$ into three parts: self-attention output $\mathbf{a}^L_{-1}$, FFN output $\mathbf{m}^L_{-1}$ and hidden states at the previous layer $\mathbf{h}^{L-1}_{-1}$ (Line 1 to Line 2). Then $\mathbf{h}^{L-1}_{-1}$ can be further decomposed through layers until we get the embedding vector of input $\mathbf{h}^{0}_{-1}$ (Line 2 to Line 3). Ultimately, we replace $\mathbf{m}^l_{-1}$ with $\mathbf{W}_{\text{out}}^{l}\mathbf{O}^l_{-1}$ (Line 3 to Line 4). Note that we have omitted layer normalization operations in Eq.~\ref{eq:formula7} through approximate assumptions for the sake of brevity.

In Eq.~\ref{eq:formula8}, we disassemble the multiplication of $\mathbf{W}_u\mathbf{W}_{\text{out}}^{l}$ and $\mathbf{O}^l_{-1}$. The dimensionality of $\mathbf{W}_u\mathbf{W}_{\text{out}}^{l}$ is $d_m\times v$. We aim at obtaining a matrix which can indicate the contribution from each neuron to each token. Accordingly, we adopt an element-wise product with broadcasting mechanism between $\mathbf{W}_u\mathbf{W}_{\text{out}}^{l}$ and $\mathcal{T}\left(\mathbf{O}^l_{-1}\right)$, keeping the original dimensionality unchanged.

We mainly focus on the last token outputs in Eq.~\ref{eq:formula9}, Eq.~\ref{eq:formula7} and Eq.~\ref{eq:formula8}. The rationale behind our approach is that an autoregressive Transformer will generate the new token at the position of the last input token. Therefore, analyzing the last token can help us understand the principles underlying the model generation process.

\section{Implementation Details}

\subsection{Identifying Multi-Modal Neurons}
\label{subsec:identifying}
For model LLaVA~\citep{liu2023visual}, we choose the version whose base LLM is LLaMA-2-13B-Chat~\citep{touvron2023llama2} and visual encoder is ViT-L/14~\citep{radford2021learning}. Each input image is resized to (224, 224) and encoded into a sequence $[z_1, \cdots, z_p]$ of dimensionality 1024, where $p=256$. Then a projection layer transforms sequence $[z_1, \cdots, z_p]$ into image prompts $[x_1, \cdots, x_p]$ of dimensionality 5120. The image prompts will be concatenated into the textual prompts and received by LLaVA.

For model InstructBLIP~\citep{instructblip}, we choose the version that employs image encoder including ViT-g/14~\citep{Fang_2023_CVPR} and a Q-former~\citep{li2023blip2}, and adopts Vicuna-7B~\citep{vicuna2023} as the LLM. Similar to LLaVA, each image is encoded into a sequence $[z'_1, \cdots, z'_q]$, where $q=256$. And then the sequence is sent into the Q-former to get the extracted image features $[z_1, \cdots, z_p]$ of dimensionality 768, where $p=32$. Then a projection layer transforms sequence $[z_1, \cdots, z_p]$ into image prompts $[x_1, \cdots, x_p]$ of dimensionality 4096.

Model mPLUG-Owl2~\citep{ye2023mplugowl2} utilizes ViT-L/14~\citep{radford2021learning} as visual encoder and LLaMA-2-7B~\citep{touvron2023llama2} as LLM. Different from LLaVA and InstructBLIP, mPLUG-Owl2 adopts a visual abstractor after the visual encoder, which transforms image features $[z_1, \cdots, z_p]$ of dimensionality 1024 into image prompts $[x_1, \cdots, x_p]$ of dimensionality 4096.

We adopt ``Describe the image in few words.'' as query prompts in all models. Note that for better captioning results, we add a text prefix ``An image of'' after the textual prompts.

We use greedy search when generating captions for each image, which means the token with the highest probability will be selected at each step. We calculate the contribution score $s_{i, t}^l$ for each nominal token $t$ in the generated caption, and rank all contribution scores across all layers within the model by the descending order to select top neurons as multi-modal neurons.

It should be noted that while we can calculate scores for all tokens generated by the model, some tokens may not be readily describable from the image content alone. Therefore, for the purpose of a clearer explanation, our analysis focuses only on tokens corresponding to nouns. If a noun consists of multiple tokens, we select the first token as being representative of that noun. To identify all nouns in the caption, we use Stanford CoreNLP~\citep{manning2014stanford}, a tool for natural language processing in Java, by an open-source python wrapper \footnote{\url{https://github.com/Jason3900/corenlp\_client}}.

We compare our method with Multimodal Neurons~\citep{schwettmann2023multimodal}, which calculates the attribution scores to select neurons. In their method, an attribution score is obtained for each image patch and neuron. For fair comparisons in our experiments, we modify this by taking the maximum attribution score across patches for each neuron. This modification avoids unnecessary repetition while maintaining the interpretability of the neuron attributions.

Furthermore, we established a baseline approach that solely considers the activations of neurons at the last input token as contribution scores, selecting those neurons exhibiting higher levels of activation as contributory neurons.

We run the experiments on NVIDIA GTX 1080Ti, NVIDIA RTX 2080Ti and NVIDIA RTX 3090 GPUs, and it takes about 500 GPU hours.

\subsection{Tracing Focus of Neurons in Images}
\label{appendix_subsec:tracing_feature_regions_in_images}

Following previous works on feature visualization~\citep{netdissect2017, schwettmann2023multimodal}, we are curious about where neurons focus their attention. To trace focus of neurons in images, we employ a visualization approach described below.

We denote the size of input images as $d_i\times d_i$. Assuming that after passing through the image encoder, there are $p$ image tokens input into the LLM. We assume that $p$ can be square rooted. For each multi-modal neuron, we take its activations at image tokens and reshape them into a $\sqrt{p} \times \sqrt{p}$ matrix. And then we scale them to $d_i \times d_i$ by bilinear interpolation. Now the scaled activations and the input images have the same size. For each image, we first plot a heatmap by using a mean scaled activation across top-$k$ neurons and put it over the image. We then threshold the mean scaled activations above the 95\% percentile to produce a binary mask and also combine it with the original image.

Since the square root of the number of image patch tokens (i.e. $\sqrt{p}$) in InstructBLIP and mPLUG-Owl2 is irrational, we only trace focus of neurons using LLaVA.

\subsection{Multi-Modal Knowledge Editing}
\label{appendix_a_subsec:targeted_modification}
For most images, we empirically pick out the top-5 multi-modal neurons as $\mathcal{S}$, initialize $\Delta\mathbf{w}$ as $\mathbf{0}$, and set the learning rate $\alpha$ as 0.001, the iteration epochs $\epsilon$ as 1000 and the penalty weight $\beta$ as 4, respectively.

\section{More Experiment Results}
We report more experiment results and show more cases here to confirm our conclusion convincingly.

\subsection{Tracing Focus of Neurons in Images}
\label{subsec:heatmap_and_binary_mask}
We report heatmap and binary mask results of examples in Table \ref{tab:heatmap_and_binary_mask}. Each heatmap is plotted by using scaled mean activations across top-$k$ neurons, where $k=1, 10, 50, 100, 500, 1000$, and each binary mask is plotted by thresholding mean activations above the 95\% percentile, respectively.

\subsection{Textual Meanings of Neurons}
\label{appendix_subsec:finding_vision_language_skill_neurons}
Table \ref{tab:finding_vision_language_skill_neurons} shows examples of multi-modal neurons. For each concept in the caption, we report its multi-modal neurons with their corresponding top-tokens and contribution scores.

\subsection{Region Invariance of Neurons}
\label{appendix_subsec:shuffing_the_input_sequence_of_image_patches}
In Table \ref{tab:shuffling_the_input_sequence_of_image_patches}, we report some example results of captions and multi-modal neurons before and after shuffling the input sequence of image patches.

\subsection{Cross-Image Invariance of Neurons}
\label{subsec:commonality_of_vision_language_skill_neurons}
\begin{figure*}[t]
    \centering
    \includegraphics[width=\linewidth]{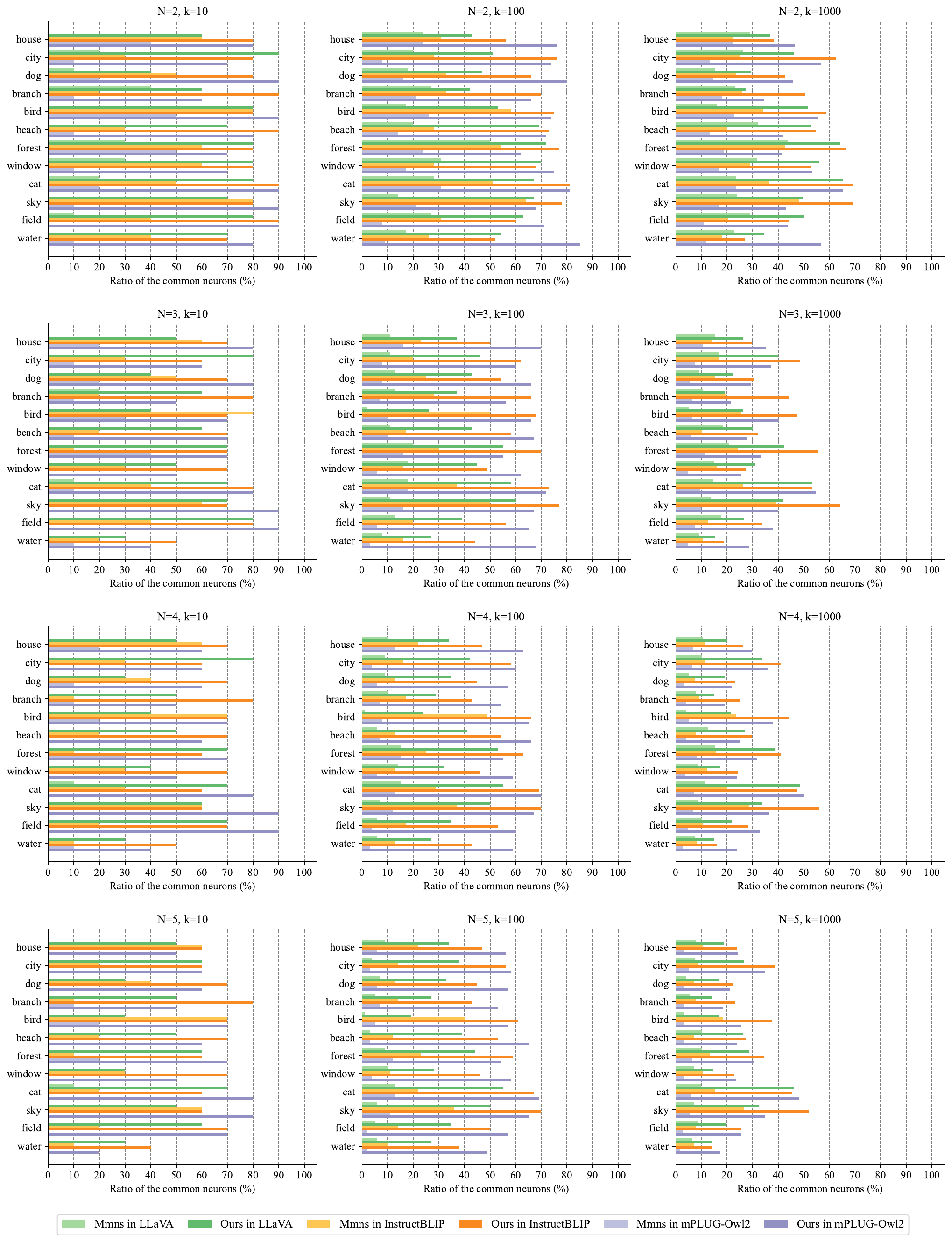}
    \caption{Ratio of the common neurons in top-$k$ neurons selected by Mmns and our method. We report $N=2, 3, 4, 5$ and $k=10, 100, 1000$ for model LLaVA, InstructBLIP and mPLUG-Owl2.}
    \label{fig:commonality}
\end{figure*}

To confirm the cross-image invariance of multi-modal neurons, in Figure \ref{fig:commonality}, we report the ratio of the common neurons in top-$k$ neurons across $N$ images that contain the same concepts, where $N=2, 3, 4, 5$ and $k=10, 100, 1000$, respectively.

\subsection{Specificity of Neurons}
\label{subsec:specificity_of_vision_language_skill_neurons}
\begin{figure*}[t]
    \centering
    \begin{varwidth}[t]{\textwidth}
    \vspace*{50pt}
    \includegraphics[width=0.16\linewidth]{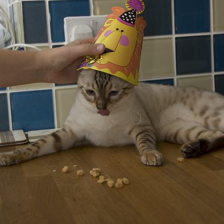} \\[5pt]
    \parbox{2.6cm}{\scriptsize LLaVA: a \textbf{cat} wearing a \textbf{birthday} \textbf{hat} and eating a \textbf{snack}, possibly a \textbf{cookie}, while sitting on a \textbf{table}.} \\[14pt]
    \parbox{2.6cm}{\scriptsize InstructBLIP: a \textbf{cat} laying on a \textbf{table} with a \textbf{birthday} \textbf{hat} on its \textbf{head}.} \\[14pt]
    \parbox{2.6cm}{\scriptsize mPLUG-Owl2: a \textbf{cat} wearing a \textbf{birthday} \textbf{hat} and a \textbf{person} feeding it.}
    \end{varwidth}
    \hspace{5pt}
    \begin{varwidth}[t]{\textwidth}
    \vspace*{0pt}
    \includegraphics[width=0.25\linewidth]{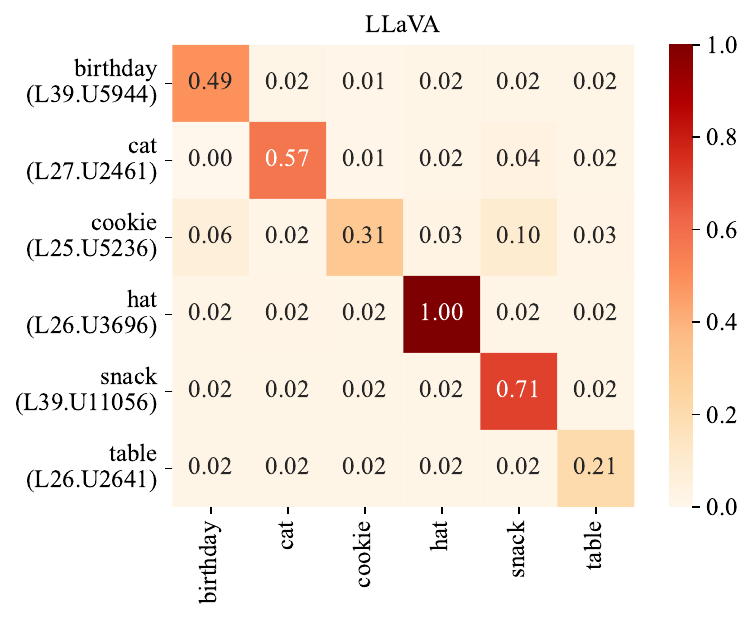} \\
    \includegraphics[width=0.25\linewidth]{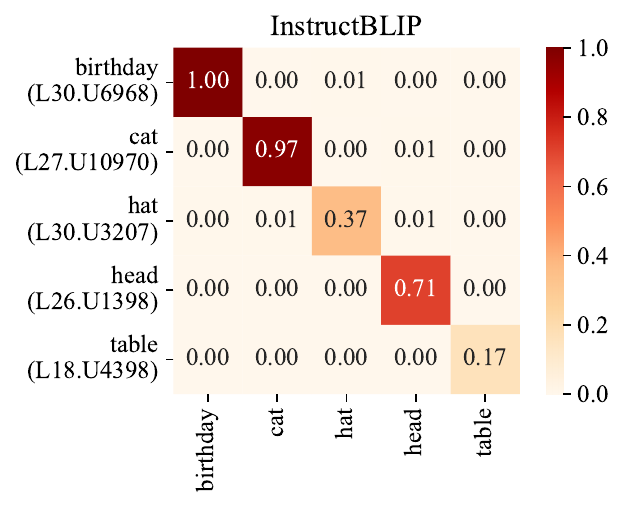} \\
    \includegraphics[width=0.25\linewidth]{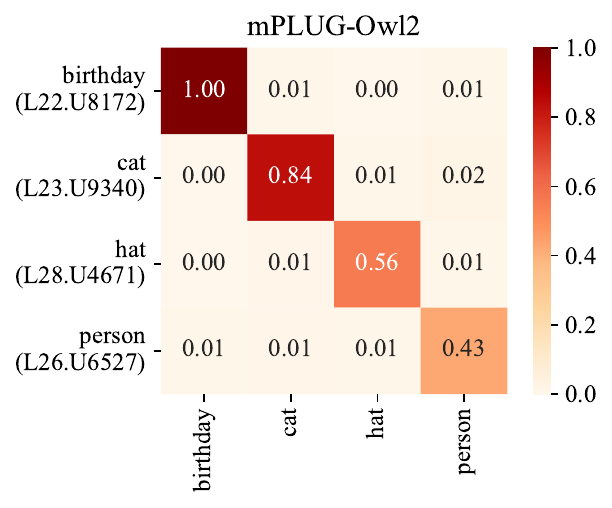}
    \end{varwidth}
    \hspace{15pt}
    \begin{varwidth}[t]{\textwidth}
    \vspace*{50pt}
    \includegraphics[width=0.16\linewidth]{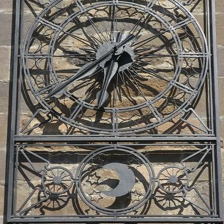} \\[5pt]
    \parbox{2.6cm}{\scriptsize LLaVA: a large \textbf{clock} on a \textbf{building}, featuring a \textbf{moon} and \textbf{sun} \textbf{design}.} \\[14pt]
    \parbox{2.6cm}{\scriptsize InstructBLIP: a \textbf{clock} on a \textbf{building} with a \textbf{metal} \textbf{frame}.} \\[14pt]
    \parbox{2.6cm}{\scriptsize mPLUG-Owl2: a \textbf{clock} with a \textbf{sun} and \textbf{moon} \textbf{design} on it.}
    \end{varwidth}
    \hspace{5pt}
    \begin{varwidth}[t]{\textwidth}
    \vspace*{0pt}
    \includegraphics[width=0.25\linewidth]{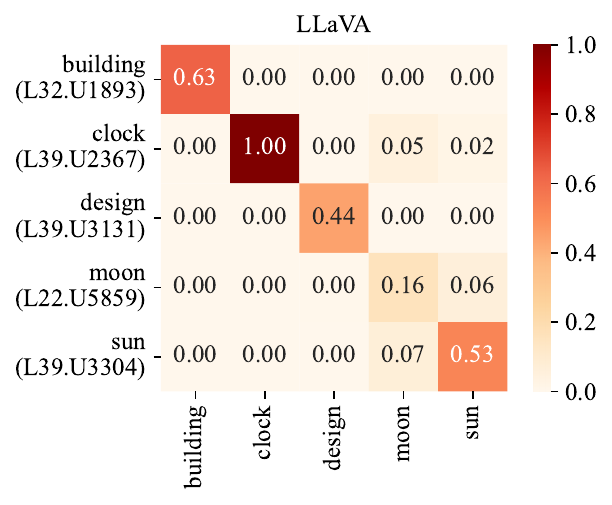} \\
    \includegraphics[width=0.25\linewidth]{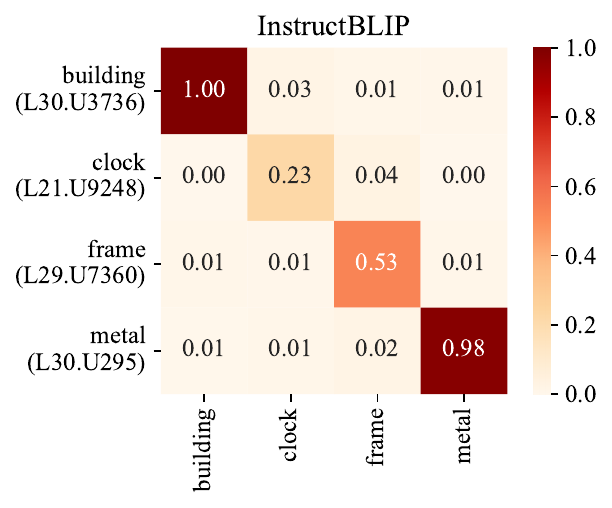} \\
    \includegraphics[width=0.25\linewidth]{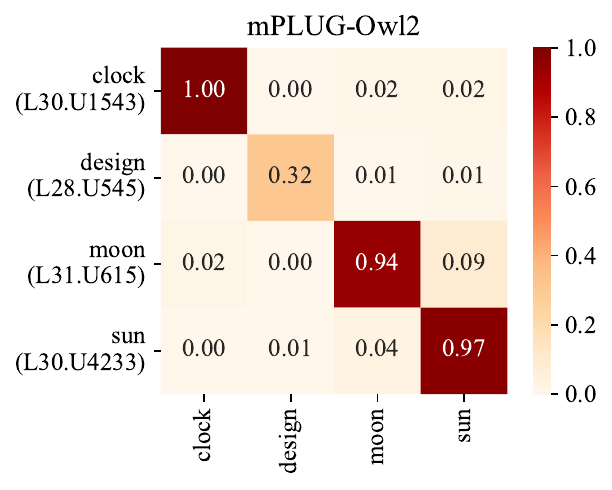}
    \end{varwidth}

    \vspace*{20pt}

    \begin{varwidth}[t]{\textwidth}
    \vspace*{30pt}
    \includegraphics[width=0.16\linewidth]{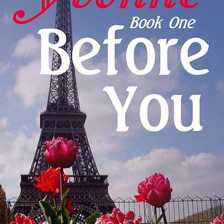} \\[5pt]
    \parbox{2.6cm}{\scriptsize LLaVA: a \textbf{book} with a \textbf{cover} featuring a \textbf{picture} of the \textbf{Eiffel} \textbf{Tower}, \textbf{flowers}, and the \textbf{words} ``Before You'' written on it.} \\[14pt]
    \parbox{2.6cm}{\scriptsize InstructBLIP: the \textbf{eiffel} \textbf{tower} with the \textbf{words}, \textbf{ivonne} \textbf{book} one before you.} \\[14pt]
    \parbox{2.6cm}{\scriptsize mPLUG-Owl2: a \textbf{book} \textbf{cover} with the \textbf{title} ``Before You'' and a \textbf{picture} of the \textbf{Eiffel} \textbf{Tower} in the \textbf{background}.}
    \end{varwidth}
    \hspace{5pt}
    \begin{varwidth}[t]{\textwidth}
    \vspace*{0pt}
    \includegraphics[width=0.25\linewidth]{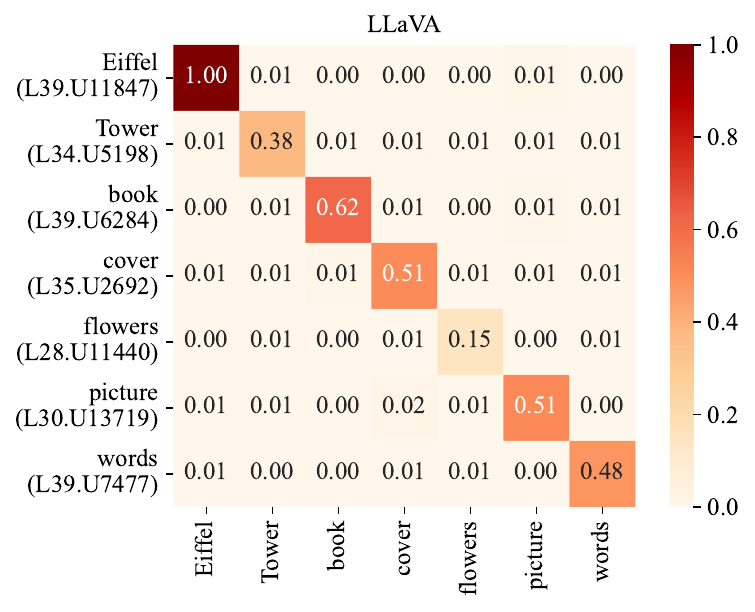} \\
    \includegraphics[width=0.25\linewidth]{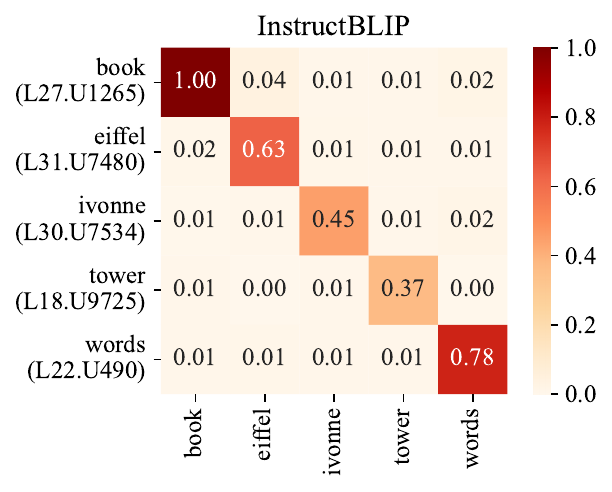} \\
    \includegraphics[width=0.25\linewidth]{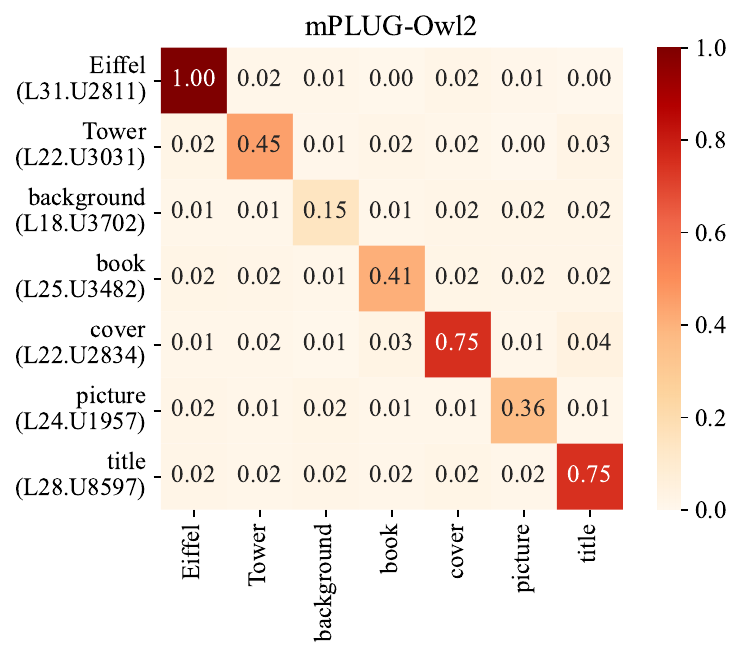}
    \end{varwidth}
    \hspace{15pt}
    \begin{varwidth}[t]{\textwidth}
    \vspace*{30pt}
    \includegraphics[width=0.16\linewidth]{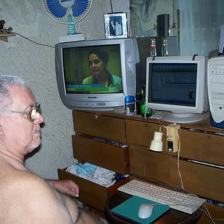} \\[5pt]
    \parbox{2.6cm}{\scriptsize LLaVA: a \textbf{man} sitting in \textbf{front} of a \textbf{computer}, with a \textbf{TV} in the \textbf{background}, and a \textbf{keyboard} on his \textbf{lap}.} \\[14pt]
    \parbox{2.6cm}{\scriptsize InstructBLIP: a \textbf{man} sitting in \textbf{front} of a \textbf{computer} \textbf{monitor}.} \\[14pt]
    \parbox{2.6cm}{\scriptsize mPLUG-Owl2: an older \textbf{man} sitting in \textbf{front} of a \textbf{television}, watching a \textbf{woman} on the \textbf{screen}.}
    \end{varwidth}
    \hspace{5pt}
    \begin{varwidth}[t]{\textwidth}
    \vspace*{0pt}
    \includegraphics[width=0.25\linewidth]{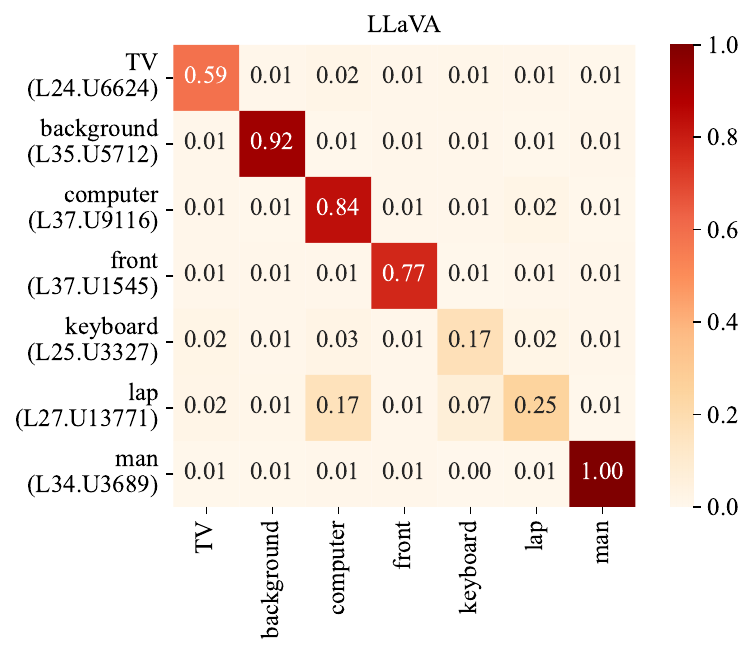} \\
    \includegraphics[width=0.25\linewidth]{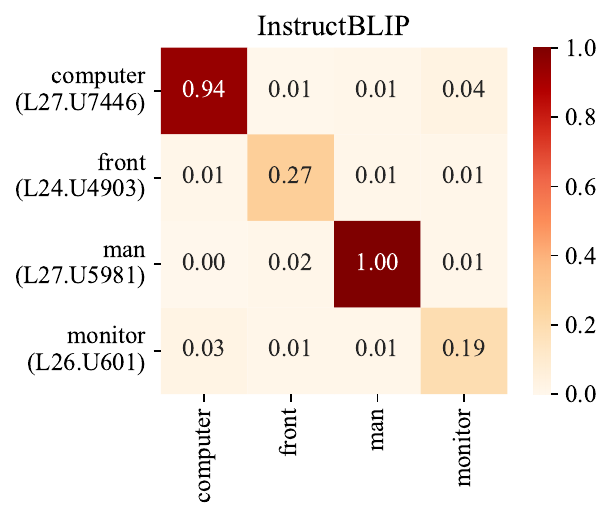} \\
    \includegraphics[width=0.25\linewidth]{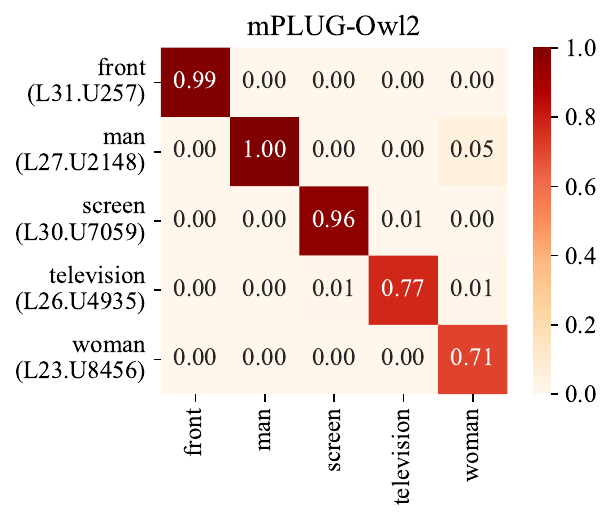}
    \end{varwidth}
    
    \caption{Heatmaps of the scores (after normalization) of multi-modal neurons corresponding to specific semantics when encoding different semantics. For each image, we report the result of the top-1 multi-modal neuron. In each heatmap, the x-axis represents concepts in the given image, and y-axis represents the top-1 neuron corresponding to each concept, respectively. Darker blocks indicate higher scores, which means higher relevance.}
    \label{fig:specificity}
\end{figure*}

To verify the specificity of multi-modal neurons, in Figure \ref{fig:specificity}, we report some examples of the heatmap of the scores of multi-modal neurons corresponding to specific concepts when encoding different concepts.

\subsection{Perturbing Multi-Modal Neurons}
\label{subsec:perturbing_vision_language_skill_neurons}
Table \ref{tab:perturbing_vision_language_skill_neurons} shows results of perturbing top-5 multi-modal neurons and 5 randomly selected neurons.

\subsection{Multi-Modal Knowledge Editing}
\label{appendix_b_subsec:targeted_modification}
Table \ref{tab:targeted_modification_examples} shows additional examples of multi-modal knowledge editing results.

\newpage
\begin{table*}[t]
    \centering
    \renewcommand\arraystretch{1.0}
\begin{tabular}{cc@{\hspace{4pt}}c@{\hspace{4pt}}c@{\hspace{4pt}}c@{\hspace{4pt}}c@{\hspace{4pt}}c}
        \toprule
        \multicolumn{7}{c}{\textbf{Image \& Original output}} \\
        \midrule
        \multicolumn{7}{c}{\hspace{0pt} \includegraphics[width=0.2\linewidth]{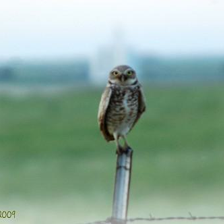} \hspace{5pt}\begin{varwidth}[f]{\textwidth}
        \vspace*{-75pt}\makecell[l]{\small \parbox{8cm}{LLaVA: a small \textbf{owl} perched on a \textbf{metal} \textbf{pole} in a grassy \textbf{field}.}}\end{varwidth}} \\
        \midrule
        \multirow{2}{*}{\bf Concept} & \multicolumn{6}{c}{\bf Heatmap \& Binary mask} \\
        & \small \bf Top-1 & \small \bf Top-10 & \small \bf Top-50 & \small \bf Top-100 & \small \bf Top-500 & \small \bf Top-1000 \\
        
        \midrule
         \multirow{5}{*}{\raisebox{-0.0\height}{owl}} & {\raisebox{-0\height}{\makecell{\includegraphics[width=0.12\textwidth]{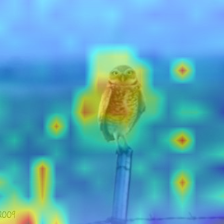}}}} & {\raisebox{-0\height}{\makecell{\includegraphics[width=0.12\textwidth]{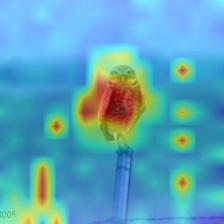}}}} & {\raisebox{-0\height}{\makecell{\includegraphics[width=0.12\textwidth]{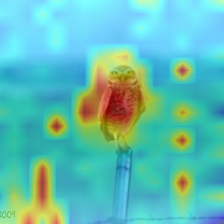}}}} & {\raisebox{-0\height}{\makecell{\includegraphics[width=0.12\textwidth]{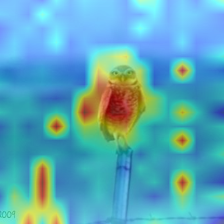}}}} & {\raisebox{-0\height}{\makecell{\includegraphics[width=0.12\textwidth]{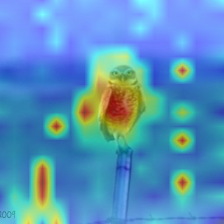}}}} & {\raisebox{-0\height}{\makecell{\includegraphics[width=0.12\textwidth]{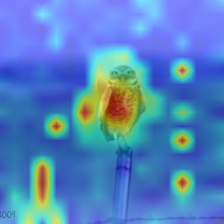}}}} \\
\bottomrule
    \end{tabular}
    \caption*{}
\end{table*}

\newpage
\begin{table*}[t]
    \centering
    \renewcommand\arraystretch{1.0}
    \begin{tabular}{cc@{\hspace{4pt}}c@{\hspace{4pt}}c@{\hspace{4pt}}c@{\hspace{4pt}}c@{\hspace{4pt}}c}
        \toprule
        \multicolumn{7}{c}{\textbf{Image \& Original output}} \\
        \midrule
        \multicolumn{7}{c}{\hspace{10pt} \includegraphics[width=0.2\linewidth]{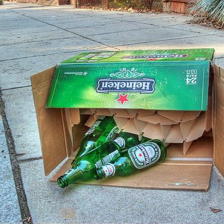} \hspace{5pt}\begin{varwidth}[f]{\textwidth}
        \vspace*{-75pt}\makecell[l]{\small \parbox{9cm}{LLaVA: a \textbf{box} filled with empty \textbf{beer} \textbf{bottles}, sitting on the \textbf{sidewalk}.}}\end{varwidth}} \\
        \midrule
        \multirow{2}{*}{\bf Concept} & \multicolumn{6}{c}{\bf Heatmap \& Binary mask} \\
        & \small \bf Top-1 & \small \bf Top-10 & \small \bf Top-50 & \small \bf Top-100 & \small \bf Top-500 & \small \bf Top-1000 \\
        
        \midrule
         \multirow{5}{*}{\raisebox{-0.0\height}{box}} & {\raisebox{-0\height}{\makecell{\includegraphics[width=0.12\textwidth]{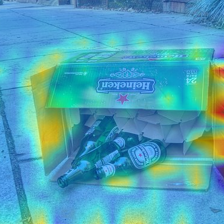}}}} & {\raisebox{-0\height}{\makecell{\includegraphics[width=0.12\textwidth]{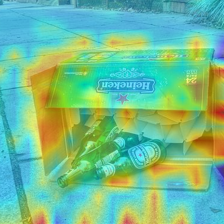}}}} & {\raisebox{-0\height}{\makecell{\includegraphics[width=0.12\textwidth]{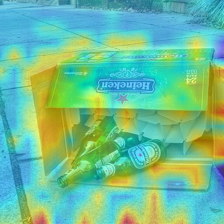}}}} & {\raisebox{-0\height}{\makecell{\includegraphics[width=0.12\textwidth]{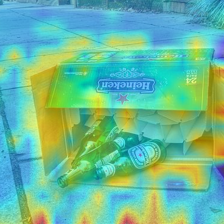}}}} & {\raisebox{-0\height}{\makecell{\includegraphics[width=0.12\textwidth]{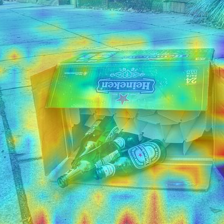}}}} & {\raisebox{-0\height}{\makecell{\includegraphics[width=0.12\textwidth]{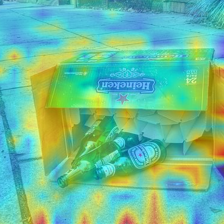}}}} \\
\bottomrule
    \end{tabular}
    \caption*{}
\end{table*}

\newpage
\begin{table*}[t]
    \centering
    \renewcommand\arraystretch{1.0}
    \begin{tabular}{cc@{\hspace{4pt}}c@{\hspace{4pt}}c@{\hspace{4pt}}c@{\hspace{4pt}}c@{\hspace{4pt}}c}
        \toprule
        \multicolumn{7}{c}{\textbf{Image \& Original output}} \\
        \midrule
        \multicolumn{7}{c}{\hspace{-8pt} \includegraphics[width=0.2\linewidth]{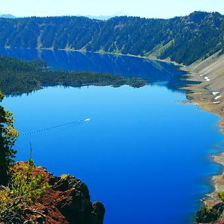} \hspace{5pt}\begin{varwidth}[f]{\textwidth}
        \vspace*{-85pt}\makecell[l]{\small \parbox{11cm}{LLaVA: a beautiful \textbf{lake} surrounded by \textbf{mountains}, with a \textbf{boat} floating on the \textbf{water}.}}\end{varwidth}} \\
        \midrule
        \multirow{2}{*}{\bf Concept} & \multicolumn{6}{c}{\bf Heatmap \& Binary mask} \\
        & \small \bf Top-1 & \small \bf Top-10 & \small \bf Top-50 & \small \bf Top-100 & \small \bf Top-500 & \small \bf Top-1000 \\
        
        \midrule
         \multirow{5}{*}{\raisebox{-0.0\height}{lake}} & {\raisebox{-0\height}{\makecell{\includegraphics[width=0.12\textwidth]{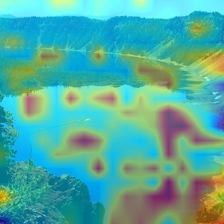}}}} & {\raisebox{-0\height}{\makecell{\includegraphics[width=0.12\textwidth]{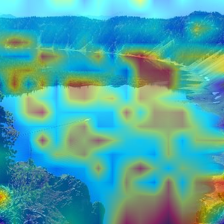}}}} & {\raisebox{-0\height}{\makecell{\includegraphics[width=0.12\textwidth]{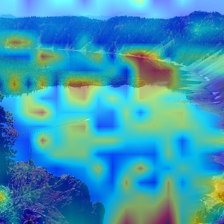}}}} & {\raisebox{-0\height}{\makecell{\includegraphics[width=0.12\textwidth]{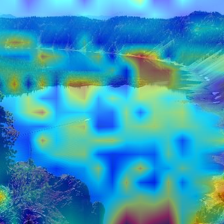}}}} & {\raisebox{-0\height}{\makecell{\includegraphics[width=0.12\textwidth]{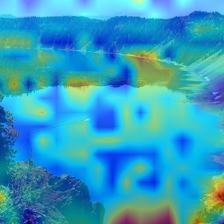}}}} & {\raisebox{-0\height}{\makecell{\includegraphics[width=0.12\textwidth]{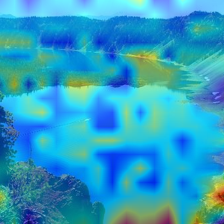}}}} \\
\bottomrule
    \end{tabular}
    \caption{Heatmap and binary mask results of example images. We plot each heatmap by using scaled mean activations across top-$k$ neurons, where $k=1, 10, 50, 100, 500, 1000$, and plot binary mask by thresholding mean activations above the 95\% percentile, respectively.}
    \label{tab:heatmap_and_binary_mask}
\end{table*}

\newpage
\begin{table*}[t]
    \small
    \centering
    \renewcommand\arraystretch{0.8}
    \begin{tabular}{c@{\hspace{5pt}}c@{\hspace{8pt}}c@{\hspace{8pt}}l@{\hspace{3pt}}p{5.0cm}@{\hspace{4pt}}c@{\hspace{2pt}}}
        \toprule
        \raisebox{-0.0\height}{\bf Image} & \raisebox{-0.0\height}{\bf Model} & \raisebox{-0.0\height}{\bf Concept} & {\bf Top neurons} & \multicolumn{1}{c}{\raisebox{-0.0\height}{\bf Top tokens}} & \raisebox{-0.0\height}{{\bf Score}} \\
        \midrule
        \multirow{1}{*}{\raisebox{-2.1\height}{\makecell{\includegraphics[width=0.2\textwidth]{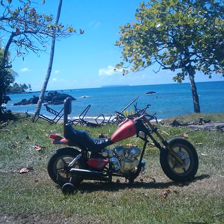} \\[4pt] \hspace{-2pt} \makecell[c]{\scriptsize \parbox{0.8cm}{\vspace*{-16pt} LLaVA: } \parbox{2.2cm}{a small red \textbf{motorcycle} parked on the \textbf{grass} near a \textbf{beach}.}} \\[16pt] \hspace{-2pt} \makecell[c]{\scriptsize \parbox{1.35cm}{\vspace*{-16pt} InstructBLIP: } \parbox{2.0cm}{a \textbf{motorcycle} parked on the \textbf{grass} near the \textbf{ocean}.}} \\[16pt] \hspace{-2pt} \makecell[c]{\scriptsize \parbox{1.54cm}{\vspace*{-16.0pt} mPLUG-Owl2: } \parbox{2.0cm}{a \textbf{motorcycle} parked on the \textbf{grass} near the \textbf{ocean}.}}}}} & \multirow{15}{*}{\raisebox{-1.0\height}{LLaVA}} & \multirow{4}{*}{\raisebox{-1.0\height}{motorcycle}} & ~\scriptsize \bf L34.U12567 & \scriptsize [`motor',`Motor',`mot',`b',`mot'] & \scriptsize 0.906 \\
        & & & ~\scriptsize \bf L33.U6828 & \scriptsize [`mot',`Mot',`mot',`motiv',`Motor'] & \scriptsize 0.850 \\
        & & & ~\scriptsize \bf L25.U11735 & \scriptsize [`motor',`tennis',`hockey',`basketball',`football'] & \scriptsize 0.641 \\
        & & & ~\scriptsize \bf L24.U5729 & \scriptsize [`vehicle',`vehicles',`aircraft',`boat',`motor'] & \scriptsize 0.591 \\
        & & & ~\scriptsize \bf L27.U11389 & \scriptsize [`mot',`motor',`Mot',`Motor',`mot'] & \scriptsize 0.533 \\
        
        \cmidrule(lr){3-6}
        
        & & \multirow{4}{*}{\raisebox{-1.0\height}{grass}} & ~\scriptsize \bf L25.U5542 & \scriptsize [`grass',`woods',`leaf',`forest',`bush'] & \scriptsize 2.039 \\
\bottomrule
    \end{tabular}
    \caption*{}
\end{table*}

\begin{table*}[t]
    \small
    \centering
    \renewcommand\arraystretch{0.8}
    \begin{tabular}{c@{\hspace{5pt}}c@{\hspace{8pt}}c@{\hspace{8pt}}l@{\hspace{3pt}}p{5.3cm}@{\hspace{4pt}}c@{\hspace{2pt}}}
        \toprule
        \raisebox{-0.0\height}{\bf Image} & \raisebox{-0.0\height}{\bf Model} & \raisebox{-0.0\height}{\bf Concept} & {\bf Top neurons} & \multicolumn{1}{c}{\raisebox{-0.0\height}{\bf Top tokens}} & \raisebox{-0.0\height}{{\bf Score}} \\
        \midrule
        \multirow{1}{*}{\raisebox{-2.4\height}{\makecell{\includegraphics[width=0.2\textwidth]{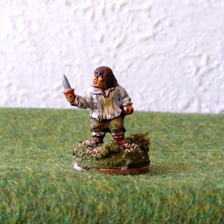} \\[4pt] \hspace{-2pt} \makecell[c]{\scriptsize \parbox{0.8cm}{\vspace*{-33.3pt} LLaVA: } \parbox{2.2cm}{a small \textbf{figurine}, possibly a \textbf{toy} or a \textbf{model}, is displayed on a green \textbf{surface}, possibly a \textbf{table} or a grassy \textbf{area}.}} \\[26pt] \hspace{-2pt} \makecell[c]{\scriptsize \parbox{1.35cm}{\vspace*{-7.8pt} InstructBLIP: } \parbox{2.0cm}{a miniature \textbf{figurine} with a \textbf{knife}.}} \\[14pt] \hspace{-2pt} \makecell[c]{\scriptsize \parbox{1.54cm}{\vspace*{-16.0pt} mPLUG-Owl2: } \parbox{1.8cm}{a small \textbf{figurine} of a \textbf{man} holding a \textbf{knife}.}}}}} & \multirow{32}{*}{\raisebox{-1.0\height}{LLaVA}} & \multirow{4}{*}{\raisebox{-1.0\height}{figurine}} & ~\scriptsize \bf L36.U8273 & \scriptsize [`figure',`Fig',`Figure',`figures',`Fig'] & \scriptsize 2.572 \\
        & & & ~\scriptsize \bf L24.U12276 & \scriptsize [`stat',`statue',`sculpt',`Stat',`stat'] & \scriptsize 1.161 \\
        & & & ~\scriptsize \bf L18.U4770 & \scriptsize [`mini',`figure',`figures',`figur',`model'] & \scriptsize 1.014 \\
        & & & ~\scriptsize \bf L38.U10971 & \scriptsize [`figure',`figures',`Figure',`Fig',`figured'] & \scriptsize 0.833 \\
        & & & ~\scriptsize \bf L38.U2195 & \scriptsize [`\foreignlanguage{russian}{Хронологија}',`Kontrola',`konn',`Audiod',`techni'] & \scriptsize 0.627 \\
\bottomrule
    \end{tabular}
    \caption*{}
\end{table*}

\begin{table*}[t]
    \small
    \centering
    \renewcommand\arraystretch{0.8}
    \begin{tabular}{c@{\hspace{5pt}}c@{\hspace{8pt}}c@{\hspace{8pt}}l@{\hspace{3pt}}p{5.3cm}@{\hspace{4pt}}c@{\hspace{2pt}}}
        \toprule
        \raisebox{-0.0\height}{\bf Image} & \raisebox{-0.0\height}{\bf Model} & \raisebox{-0.0\height}{\bf Concept} & {\bf Top neurons} & \multicolumn{1}{c}{\raisebox{-0.0\height}{\bf Top tokens}} & \raisebox{-0.0\height}{{\bf Score}} \\
        \midrule
        \multirow{1}{*}{\raisebox{-2.4\height}{\makecell{\includegraphics[width=0.2\textwidth]{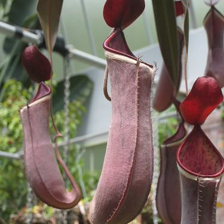} \\[4pt] \hspace{-2pt} \makecell[c]{\scriptsize \parbox{0.8cm}{\vspace*{-25.4pt} LLaVA: } \parbox{2.5cm}{a \textbf{plant} with red \textbf{flowers} hanging from it, possibly a  Venus \textbf{flytrap}, is displayed in a \textbf{greenhouse}.}} \\[22pt] \hspace{-2pt} \makecell[c]{\scriptsize \parbox{1.35cm}{\vspace*{-9.2pt} InstructBLIP: } \parbox{2.0cm}{a \textbf{pitcher} \textbf{plant} hanging from a \textbf{tree}.}} \\[12pt] \hspace{-2pt} \makecell[c]{\scriptsize \parbox{1.54cm}{\vspace*{-24.0pt} mPLUG-Owl2: } \parbox{1.8cm}{a \textbf{pitcher} \textbf{plant} hanging from a \textbf{ceiling} in a \textbf{greenhouse}.}}}}} & \multirow{20}{*}{\raisebox{-1.0\height}{LLaVA}} & \multirow{4}{*}{\raisebox{-1.0\height}{plant}} & ~\scriptsize \bf L27.U8060 & \scriptsize [`plant',`Plant',`plant',`plants',`planta'] & \scriptsize 1.087 \\
        & & & ~\scriptsize \bf L29.U9056 & \scriptsize [`shr',`bush',`Bush',`plant',`plants'] & \scriptsize 0.962 \\
        & & & ~\scriptsize \bf L28.U11440 & \scriptsize [`flow',`blo',`Flow',`blo',`Flow'] & \scriptsize 0.621 \\
        & & & ~\scriptsize \bf L27.U498 & \scriptsize [`branch',`Branch',`branches',`branch',`bush'] & \scriptsize 0.600 \\
        & & & ~\scriptsize \bf L25.U11504 & \scriptsize [`roots',`root',`Root',`root',`leaves'] & \scriptsize 0.502 \\
\bottomrule
    \end{tabular}
    \caption{Multi-modal neurons with their corresponding top tokens and their contribution scores. For each concept in the caption, we report the top-5 neurons with the top-5 highest probability of tokens.}
    \label{tab:finding_vision_language_skill_neurons}
\end{table*}

\newpage
\begin{table*}[t]
    \centering
    \renewcommand\arraystretch{0.7}
    \begin{tabular}{cp{5.6cm}p{5.5cm}}
        \toprule
        {\bf Image} & \multicolumn{1}{c}{\bf Original} & \multicolumn{1}{c}{\bf Shuffled} \\
        \midrule
        
        \raisebox{0\height}{\makecell{\includegraphics[width=0.2\textwidth]{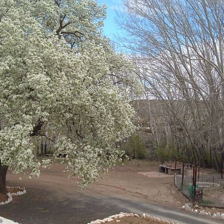}}} & \raisebox{-0pt}{\makecell[l]{\small \parbox{5.5cm}{a \textbf{tree} with white \textbf{flowers} in a \textbf{field}, surrounded by a \textbf{dirt} \textbf{road} and a \textbf{fence}. \\[8pt] \small \bf tree: \\[2pt] \scriptsize [L28.U9085, L36.U1422, L22.U171, L27.U8824] \\[2pt] \small \bf flowers: \\[2pt] \scriptsize [L28.U11440, L20.U8129, L27.U13027, L27.U498] \\[2pt] \small \bf field: \\[2pt] \scriptsize [L34.U12955, L28.U1085, L25.U5542, L39.U7153] \\[2pt] \small \bf dirt: \\[2pt] \scriptsize [L39.U8730, L31.U526, L39.U212, L35.U1480] \\[2pt] \small \bf road: \\[2pt] \scriptsize [L39.U8637, L26.U1456, L37.U12619, L29.U224] \\[2pt] \small \bf fence: \\[2pt] \scriptsize [L27.U12313, L38.U5969, L37.U2453, L39.U212]}}} & \raisebox{-0pt}{\makecell[l]{\small \parbox{5.5cm}{a \textbf{tree} with white \textbf{flowers} in a \textbf{field}, surrounded by a \textbf{dirt} \textbf{road} and a \textbf{fence}. \\[8pt] \small \bf tree: \\[2pt] \scriptsize [L28.U9085, L36.U1422, L22.U171, L27.U8824] \\[2pt] \small \bf flowers: \\[2pt] \scriptsize [L28.U11440, L20.U8129, L27.U13027, L27.U498] \\[2pt] \small \bf field: \\[2pt] \scriptsize [L34.U12955, L28.U1085, L25.U5542, L39.U7153] \\[2pt] \small \bf dirt: \\[2pt] \scriptsize [L31.U526, L39.U8730, L39.U212, L35.U1480] \\[2pt] \small \bf road: \\[2pt] \scriptsize [L39.U8637, L26.U1456, L37.U12619, L29.U224] \\[2pt] \small \bf fence: \\[2pt] \scriptsize [L27.U12313, L38.U5969, L37.U2453, L39.U212]}}} \\
\bottomrule
    \end{tabular}
    \caption{Example results of captions and multi-modal neurons before and after shuffling the input sequence of image patches, respectively. We just record the concepts that appear both in original and shuffled captions from LLaVA, and for each concept, we report its top-4 multi-modal neurons.}
    \label{tab:shuffling_the_input_sequence_of_image_patches}
\end{table*}

\newpage

\newpage

\newpage
\begin{table*}[t]
    \centering
    \renewcommand\arraystretch{0.9}
    \begin{tabular}{ccp{9.2cm}}
        \toprule
        \bf Image & \bf Concept & \multicolumn{1}{c}{\bf Perturbed model output} \\
        \midrule
        \multirow{1}{*}{\raisebox{-1.05\height}{\makecell{\includegraphics[width=0.2\textwidth]{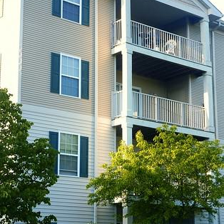} \\ \hspace{-2pt} \makecell[l]{\small \parbox{1.02cm}{\vspace*{-31.8pt} LLaVA: } \parbox{2.4cm}{a tall \textbf{apartment} \textbf{building} with \textbf{balconies} and a \textbf{tree} in the \textbf{background}.}}}}} & {apartment} & a multilevelishiigledishiigledishiigledishiigledishi... \\
        \cmidrule(lr){2-3}
        & \multirow{2}{*}{building} & a white and blue building with a balcony and a tree in the background. \\
        \cmidrule(lr){2-3}
        & \multirow{2}{*}{balconies} & a building with eradicated trees in the background, with eradicated trees on eradicated trees on 2200. \\
        \cmidrule(lr){2-3}
        & {tree} & a white building with a balcony and a chair on it. \\
        \cmidrule(lr){2-3}
        & \multirow{2}{*}{background} & a tall apartment building with balconies and a tree in front of it. \\
        \cmidrule(lr){2-3}
        & \multirow{2}{*}{\textit{random}} & a tall apartment building with balconies and a tree in the background. \\
        
        \midrule

        \multirow{1}{*}{\raisebox{-1.15\height}{\makecell{\includegraphics[width=0.2\textwidth]{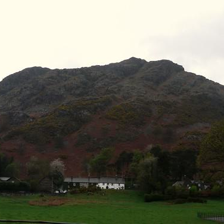} \\ \hspace{-2pt} \makecell[l]{\small \parbox{1.02cm}{\vspace*{-39.8pt} LLaVA: } \parbox{2.5cm}{a mountainous \textbf{landscape} with a \textbf{village} in the \textbf{valley}, featuring a grassy \textbf{field} and a \textbf{road}.}}}}} & \multirow{2}{*}{landscape} & a mountain range with a village in the valley, surrounded by a green field. \\
        \cmidrule(lr){2-3}
        & \multirow{2}{*}{village} & a mountain with a small town or village located at its base, surrounded by a lush green field. \\
        \cmidrule(lr){2-3}
        & \multirow{2}{*}{valley} & a mountain with a lush green field in the background, surrounded by a village. \\
        \cmidrule(lr){2-3}
        & \multirow{2}{*}{field} & a mountain with a village in the valley below, surrounded by a lush green countryside. \\
        \cmidrule(lr){2-3}
        & \multirow{2}{*}{road} & a mountainous landscape with a village in the valley, surrounded by a lush green field. \\
        \cmidrule(lr){2-3}
        & \multirow{2}{*}{\textit{random}} & a mountainous landscape with a village in the valley, featuring a grassy field and a road. \\

        \midrule

        \multirow{1}{*}{\raisebox{-1\height}{\makecell{\includegraphics[width=0.2\textwidth]{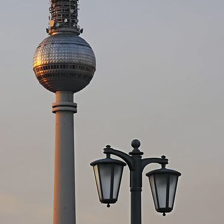} \\ \hspace{-2pt} \makecell[l]{\small \parbox{1.02cm}{\vspace*{-31.8pt} LLaVA: } \parbox{2.4cm}{a large \textbf{tower} with a \textbf{ball} on top, standing next to a \textbf{street} \textbf{light}.}}}}} & \multirow{2}{*}{tower} & a large, white building with a light on a black background, with a lighted street lamp in the foreground. \\
        \cmidrule(lr){2-3}
        & \multirow{2}{*}{ball} & a tall building with a light on top, possibly a light tower, and a street light nearby. \\
        \cmidrule(lr){2-3}
        & \multirow{2}{*}{street} & a large white building with a clock tower, a street light, and a city skyline in the background. \\
        \cmidrule(lr){2-3}
        & \multirow{3}{*}{light} & a tall tower with a large sphere on top, a building with a large sphere on top, and a building with a large sphere on top, all located in a city. \\
        \cmidrule(lr){2-3}
        & {\textit{random}} & a large tower with a ball on top, standing next to a street light. \\
        
        \midrule

        \multirow{1}{*}{\raisebox{-0.98\height}{\makecell{\includegraphics[width=0.2\textwidth]{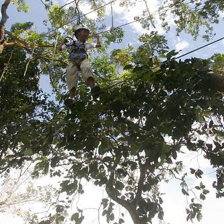} \\ \hspace{-2pt} \makecell[l]{\small \parbox{1.02cm}{\vspace*{-21.8pt} LLaVA: } \parbox{2.6cm}{a \textbf{man} hanging from a \textbf{tree} \textbf{branch} while wearing a \textbf{hat}.}}}}} & \multirow{2}{*}{man} & a person hanging from a tree branch, possibly participating in a tree-climbing activity or adventure. \\
        \cmidrule(lr){2-3}
        & \multirow{2}{*}{tree} & a man hanging from a tree, possibly on a rope or a swing, with a sky background. \\
        \cmidrule(lr){2-3}
        & \multirow{2}{*}{branch} & a man hanging from a suspended rope in a tree in the outdoors. \\
        \cmidrule(lr){2-3}
        & \multirow{2}{*}{hat} & a person hanging from a tree branch, possibly participating in a ropes course or a similar activity. \\
        \cmidrule(lr){2-3}
        & {\textit{random}} & a man hanging from a tree branch while wearing a hat. \\
        \bottomrule
    \end{tabular}
    \caption{Perturbation results of example images. For each concept in the image, we pick out top-5 multi-modal neurons and each add a Gaussian noise to perturb them. We also report results of perturbing 5 randomly selected neurons for comparison.}
    \label{tab:perturbing_vision_language_skill_neurons}
\end{table*}

\newpage
\begin{table*}[t]
    \centering
    \renewcommand\arraystretch{0.8}
    \begin{tabular}{cccp{8.3cm}}
        \toprule
        \bf Image & \bf Source & \bf Target & \multicolumn{1}{c}{\bf Edited model output} \\
        \midrule
        \multirow{1}{*}{\raisebox{-1.55\height}{\makecell{\includegraphics[width=0.2\textwidth]{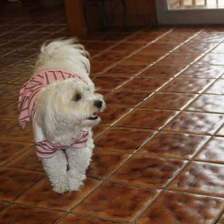} \\ \hspace{-2pt} \makecell[l]{\small \parbox{1.02cm}{\vspace*{-20pt} LLaVA: } \parbox{2.1cm}{a small white \textbf{dog} standing on a tiled \textbf{floor}.}}}}} & \multirow{6}{*}{\raisebox{-0.9\height}{dog}} & {mouse} & \small a \textbf{mouse} in a pink dress, standing on a tiled floor. \\
        \cmidrule(lr){3-4}
        & & \multirow{2}{*}{bag} & \small a white \textbf{bag} \textbf{bag} \textbf{bag}, or a white \textbf{bag} with a pink stripe, is standing on a tiled floor. \\
        \cmidrule(lr){3-4}
        & & \multirow{2}{*}{dinosaur} & \small a small white sauce \textbf{dinosaur} (dino) or a small white sauce-covered \textbf{dinosaur} toy is standing on a tiled floor. \\
        \cmidrule(lr){3-4}
        & & {baby} & \small a \textbf{baby} white \textbf{baby}, wearing a pink dress, standing on a tiled floor. \\
        
        \cmidrule(lr){2-4}
        
        & \multirow{10}{*}{\raisebox{-0.3\height}{floor}} & \multirow{2}{*}{bed} & \small a small white dog wearing a pink dress, standing on \textbf{bed} in a bedroom. \\
        \cmidrule(lr){3-4}
        & & \multirow{2}{*}{chair} & \small a small white dog wearing a pink dress, standing on \textbf{chair} in a \textbf{chair}, and looking at the camera. \\
        \cmidrule(lr){3-4}
        & & {tree} & \small a small white dog standing on a \textbf{tree} \textbf{tree} in a \textbf{tree}-lined area. \\
        \cmidrule(lr){3-4}
        & & \multirow{2}{*}{beach} & \small a small white dog, possibly a Maltese, wearing a pink dress and standing on the \textbf{beach}. \\
        \cmidrule(lr){3-4}
        & & {table} & \small a small white dog wearing a pink dress, standing on a \textbf{table}. \\
        \cmidrule(lr){3-4}
        & & {ship} & \small a small white dog standing on a \textbf{ship} shipment. \\
        
        \midrule
        
        \multirow{1}{*}{\raisebox{-3.0\height}{\makecell{\includegraphics[width=0.2\textwidth]{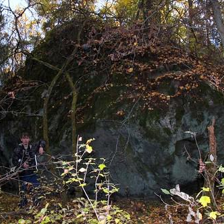} \\ \hspace{-2pt} \makecell[l]{\small \parbox{1.02cm}{\vspace*{-32pt} LLaVA: } \parbox{2.1cm}{a \textbf{man} standing in a \textbf{forest}, surrounded by \textbf{trees} and a large \textbf{rock}.}}}}} & \multirow{6}{*}{\raisebox{-0.8\height}{man}} & {woman} & \small a \textbf{woman} standing in a forest, surrounded by trees and rocks. \\
        \cmidrule(lr){3-4}
        & & {baby} & \small a \textbf{baby} in a forest, surrounded by trees and rocks. \\
        \cmidrule(lr){3-4}
        & & {girl} & \small a \textbf{girl} standing in a forest, surrounded by trees and rocks. \\
        \cmidrule(lr){3-4}
        & & \multirow{2}{*}{stone} & \small a \textbf{stone} wall or rock formation in a forest, with a \textbf{stone} cave or a rocky outcropping, and a \textbf{stone} wall with a \textbf{stone} door. \\

        \cmidrule(lr){2-4}
        
        & \multirow{12}{*}{\raisebox{-1\height}{forest}} & \multirow{2}{*}{mountain} & \small a man standing in a mountainous \textbf{mountain} area, surrounded by trees and rocks. \\
        \cmidrule(lr){3-4}
        & & {garden} & \small a man standing in a \textbf{garden} with a large rock and a tree. \\
        \cmidrule(lr){3-4}
        & & \multirow{2}{*}{water} & \small a man standing in a \textbf{waterlogged} area, surrounded by a \textbf{waterfall} and a rocky cliff. \\
        \cmidrule(lr){3-4}
        & & \multirow{2}{*}{city} & \small a man standing in a \textbf{city} park, surrounded by trees and a large rock formation. \\
        \cmidrule(lr){3-4}
        & & \multirow{2}{*}{desert} & \small a man standing in a deserted \textbf{desert} area, surrounded by trees and a large rock. \\
        \cmidrule(lr){3-4}
        & & \multirow{2}{*}{hall} & \small a man standing in a \textbf{hallway} of a cave, surrounded by rocks and trees. \\

        \cmidrule(lr){2-4}
        
        & \multirow{6}{*}{\raisebox{-1\height}{trees}} & \multirow{2}{*}{wild} & \small a man standing in front of a \textbf{wild}, \textbf{wild} rock formation, surrounded by wildlife and a forest trees. \\
        \cmidrule(lr){3-4}
        & & \multirow{2}{*}{flowers} & \small a man standing in a forest, surrounded by \textbf{flowers} and \textbf{flowers} in the background. \\
        \cmidrule(lr){3-4}
        & & \multirow{2}{*}{cloud} & \small a man standing in front of a \textbf{cloudy} sky, surrounded by a forest trees and \textbf{cloudy} sky. \\
        
        \cmidrule(lr){2-4}
        
        & \multirow{11}{*}{\raisebox{-1\height}{rock}} & {house} & \small a man standing in a forest, surrounded by trees and a large \textbf{house}. \\
        \cmidrule(lr){3-4}
        & & \multirow{2}{*}{tower} & \small a man standing in a forest, surrounded by trees and a towering \textbf{tower} towering over him. \\
        \cmidrule(lr){3-4}
        & & {building} & \small a man standing in a forest, surrounded by trees and a large \textbf{building}. \\
        \cmidrule(lr){3-4}
        & & \multirow{2}{*}{ball} & \small a man standing in a forest, surrounded by trees and a large \textbf{ball} of moss. \\
        \cmidrule(lr){3-4}
        & & \multirow{2}{*}{bear} & \small a man standing in a forest, surrounded by trees and bushes, with a large \textbf{bear} \textbf{bear} nearby. \\
        \cmidrule(lr){3-4}
        & & \multirow{2}{*}{human} & \small a man standing in a forest, surrounded by trees and a large \textbf{human}-like figure made of leaves. \\
        \bottomrule
    \end{tabular}
    \caption*{}
\end{table*}

\begin{table*}[t]
    \centering
    \renewcommand\arraystretch{0.8}
    \begin{tabular}{cccp{8.3cm}}
        \toprule
        \bf Image & \bf Source & \bf Target & \multicolumn{1}{c}{\bf Edited model output} \\
        \midrule
        \multirow{1}{*}{\raisebox{-2.05\height}{\makecell{\includegraphics[width=0.2\textwidth]{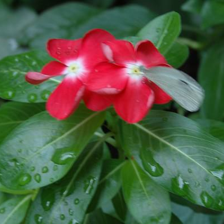} \\ \hspace{-2pt} \makecell[l]{\small \parbox{1.02cm}{\vspace*{-32pt} LLaVA: } \parbox{2.1cm}{a red \textbf{flower} with a \textbf{butterfly} on it, sitting on a leafy \textbf{plant}.}}}}} & \multirow{8}{*}{\raisebox{-0.8\height}{flower}} & {head} & \small a \textbf{head} of a red and white plant with a butterfly on it. \\
        \cmidrule(lr){3-4}
        & & {book} & \small a red \textbf{book} with a white butterfly on it, placed on a bookshelf. \\
        \cmidrule(lr){3-4}
        & & {road} & \small a red \textbf{roadside} plant with a white butterfly on it. \\
        \cmidrule(lr){3-4}
        & & {vehicle} & \small a red \textbf{vehicle} with a white butterfly on top of it. \\
        \cmidrule(lr){3-4}
        & & {stream} & \small a red and white flowering plant with a butterfly on \textbf{stream}. \\
        \cmidrule(lr){3-4}
        & & \multirow{2}{*}{salt} & \small a \textbf{salt} and pepper plant with a \textbf{salt} and pepper \textbf{salt} shaker on top of it. \\

        \cmidrule(lr){2-4}
        
        & \multirow{4}{*}{\raisebox{-1.2\height}{butterfly}} & {insect} & \small a flower with \textbf{insect} wings on it, sitting on a plant. \\
        \cmidrule(lr){3-4}
        & & \multirow{2}{*}{light} & \small a \textbf{light} red flower with white petals, sitting on a leafy plant, and surrounded by \textbf{light} rain. \\
        \cmidrule(lr){3-4}
        & & {rain} & \small a red flower with \textbf{rain} drops on it, sitting on a leafy plant. \\

        \cmidrule(lr){2-4}
        
        & \multirow{11}{*}{\raisebox{-0.8\height}{plant}} & {tree} & \small a \textbf{tree} \textbf{tree} with a red flower and a butterfly on it. \\
        \cmidrule(lr){3-4}
        & & {wall} & \small a red flower with a butterfly on it, sitting on a \textbf{wall}. \\
        \cmidrule(lr){3-4}
        & & \multirow{2}{*}{ground} & \small a red flower with a butterfly on it, sitting on a \textbf{ground} with green leaves. \\
        \cmidrule(lr){3-4}
        & & {bowl} & \small a red flower with a butterfly on it, sitting on a \textbf{bowl} \textbf{bowl} of water. \\
        \cmidrule(lr){3-4}
        & & \multirow{2}{*}{tower} & \small a towering \textbf{tower} of red flowers planted in a towering \textbf{tower} of green towering \textbf{tower}. \\
        \cmidrule(lr){3-4}
        & & {park} & \small a red flower with a butterfly on it, sitting on a \textbf{park} bench. \\
        \cmidrule(lr){3-4}
        & & {cup} & \small a red flower with a butterfly on it, sitting on a \textbf{cup}. \\
        
        \midrule
        
        \multirow{1}{*}{\raisebox{-2.4\height}{\makecell{\includegraphics[width=0.2\textwidth]{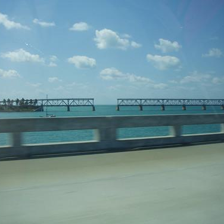} \\ \hspace{-2pt} \makecell[l]{\small \parbox{1.02cm}{\vspace*{-32pt} LLaVA: } \parbox{2.2cm}{a \textbf{bridge} over a body of \textbf{water}, with a \textbf{boat} passing underneath it.}}}}} & \multirow{10}{*}{\raisebox{-0.8\height}{bridge}} & \multirow{2}{*}{floor} & \small a \textbf{floor} with a view of a \textbf{floor}, a car, and a \textbf{floor} with a view of a car. \\
        \cmidrule(lr){3-4}
        & & \multirow{2}{*}{tree} & \small a \textbf{tree}-lined road with a \textbf{tree} in the foreground, a \textbf{tree} in the middle, and a \textbf{tree} in the background. \\
        \cmidrule(lr){3-4}
        & & \multirow{2}{*}{building} & \small a \textbf{building} with a large \textbf{building} in the background, a boat on the water, and a \textbf{building} on the water. \\
        \cmidrule(lr){3-4}
        & & \multirow{2}{*}{bed} & \small a \textbf{bed} with a view of a bedroom and a bedroom window with a view of a \textbf{bed}. \\
        \cmidrule(lr){3-4}
        & & {hill} & \small a hillside overlooking a \textbf{hill} with a hillside road and a hillside \textbf{hill}. \\

        \cmidrule(lr){2-4}
        
        & \multirow{9}{*}{\raisebox{-0.8\height}{water}} & \multirow{2}{*}{beach} & \small a bridge over a \textbf{beach}, with a boat in the background and a car driving on the bridge. \\
        \cmidrule(lr){3-4}
        & & \multirow{2}{*}{heaven} & \small a bridge over heavenly blue \textbf{heaven}, with a boat passing underneath it. \\
        \cmidrule(lr){3-4}
        & & {fire} & \small a bridge over a large body of \textbf{fire}, with a boat in the background. \\
        \cmidrule(lr){3-4}
        & & \multirow{2}{*}{snow} & \small a bridge over \textbf{snowy} mountains, with a boat traveling underneath it. \\
        \cmidrule(lr){3-4}
        & & {city} & \small a bridge over a large body of \textbf{city}, with a boat visible in the distance. \\

        \cmidrule(lr){2-4}
        
        & \multirow{8}{*}{\raisebox{-1\height}{boat}} & \multirow{2}{*}{plane} & \small a bridge over a body of water, with a \textbf{plane} flying in the background. \\
        \cmidrule(lr){3-4}
        & & \multirow{2}{*}{vehicle} & \small a bridge over a body of water, with a \textbf{vehicle} driving on it, and a \textbf{vehicle} on the other side of the bridge. \\
        \cmidrule(lr){3-4}
        & & {horse} & \small a \textbf{horse}-drawn carriage traveling on a bridge over a body of water. \\
        \cmidrule(lr){3-4}
        & & {moon} & \small a bridge over a body of water, with a \textbf{moon} in the background. \\
        \cmidrule(lr){3-4}
        & & \multirow{2}{*}{sun} & \small a \textbf{sunny} day with a bridge over a body of water, with a \textbf{sunny} sky in the background. \\
        \bottomrule
    \end{tabular}
    \caption{Knowledge editing results of example images. For each source concept in the image, we artificially transform it to other target concepts. Target concepts are in bold in the edited model output.}
    \label{tab:targeted_modification_examples}
\end{table*}

\end{document}